# Experimenting with an Evaluation Framework for Imbalanced Data Learning (EFIDL)

By

Chenyu Li MS [1], Xia Jiang PhD[1]

**[1]University of Pittsburgh School of Medicine Department of Biomedical Informatics, Pittsburgh, PA**

Corresponding Author:
Xia Jiang xij6@pitt.edu
5607 Baum Blvd, Pittsburgh, PA 15206

# ABSTRACT

## Introduction

Data imbalance is one of the crucial issues in big data analysis with fewer labels. For example, in real-world healthcare data, spam detection labels, and financial fraud detection datasets. Many data balance methods were introduced to improve machine learning algorithms' performance. Research claims SMOTE and SMOTE-based data-augmentation (generate new data points) methods could improve algorithm performance. However, we found in many online tutorials, the valuation methods were applied based on synthesized datasets that introduced bias into the evaluation, and the performance got a false improvement.
In this study, we proposed, a new evaluation framework for imbalanced data learning methods. We have experimented on five data balance methods and whether the algorithms' performance will improve or not.

## Methods

We collected 8 imbalanced healthcare datasets with different imbalanced rates from different domains. Applied 6 data augmentation methods with 11 machine learning methods testing if the data augmentation will help with improving machine learning performance. We compared the traditional data augmentation evaluation methods with our proposed cross-validation evaluation framework

## Results

Using traditional data augmentation evaluation meta hods will give a false impression of improving the performance. However, our proposed evaluation method shows data augmentation has limited ability to improve the results.

## Conclusion

EFIDL is more suitable for evaluating the prediction performance of an ML method when data are augmented. Using an unsuitable evaluation framework will give false results. Future researchers should consider the evaluation framework we proposed when dealing with augmented datasets. Our experiments showed data augmentation does not help improve ML prediction performance.

# INTRODUCTION

Data imbalance refers to a phenomenon where the number of instances of different classes in a dataset is not equally distributed. In machine learning, data imbalance can lead to a bias in the training and prediction of models, as the algorithm will tend to favor the majority class.[1] This can result in poor performance in the minority class, which is often the class of interest in many applications. Data imbalance is a common issue in real-world datasets, particularly in areas such as medical diagnosis, fraud detection, and rare event prediction.[2] Therefore, addressing data imbalance is a crucial step in building accurate and reliable machine learning models.

Data imbalance occurs when the number of cases of the targeted classes is substantially unequal. Using imbalanced data learning in breast cancer metastasis prediction as an example, breast cancer metastasis spreads cancer cells from the primary site to other body parts. In a binary classification task, we would like to predict whether a patient will develop breast cancer metastasis within five years after the initial treatment. The dataset used in the analysis will have two possible values: "yes," representing the cases when breast cancer metastasis occurred, and "no," representing the cases when it did not occur. Due to the nature of the disease, most patients will not develop metastasis within five years, resulting in a skewed distribution of the two classes. This will pose a challenge in developing accurate and reliable models for predicting breast cancer metastasis. [3,4].

Researchers have used different approaches to handle data imbalance; one of the most common approaches is data augmentation. Data augmentation involves generating new data points from existing ones to artificially increase the number of instances of the minority class. This can be done by adding slightly changed copies of previously existing data or creating synthetic data. By increasing the quantity of data, data augmentation helps to prevent overfitting and improve the performance of machine learning models. Some of the popular data augmentation techniques include:

1. Random Oversampling (ROS): This technique increases the number of instances of the minority class by randomly replicating existing cases. ROS is simple and easy to implement, but it can lead to overfitting, especially when the minority class is small.[4]
2. Synthetic Minority Over-sampling Technique (SMOTE): SMOTE generates synthetic data points by interpolating between existing minority class instances. This technique aims to balance the class distribution while preserving the original distribution of the feature space. However, SMOTE can lead to overgeneralization and overfitting if the number of synthetic instances is too high[5].
3. Adaptive Synthetic Sampling (ADASYN): ADASYN generates synthetic data points based on the density of the existing data. It adapts the number of synthetic instances to the density of the minority class, generating more instances where the density is low and fewer where the density is high. The advantage of ADASYN over SMOTE is that it tends to focus on the more challenging samples, which can improve the classification performance. However, ADASYN may generate synthetic samples not representative of the true minority class distribution.[6]

4. Other techniques: Other techniques are also used, such as Random under-sampling (RUS), Tomek links, and cost-sensitive learning.[1] Each of them has its advantages and limitations depending on the data.

These techniques are widely used in balancing the class distribution and improving model performance[7-9].

Researcher designed data augmentation methods for enhancing the performance and results of machine learning models by adding fresh and unique training samples. Some researchers suggest that using data augmentation approaches can make machine learning models more robust by simulating variability in real-world data. [10-12]However, criticisms also exist. Stephen E. Fienberg ridiculed resampling by staging, "You are attempting to get something for nothing. You utilize the same numbers again until you get an answer that cannot be obtained in any other manner. In order to do this, you must make an assumption, which you may come to regret in the future [13]. Therefore, there still exists uncertainty as to whether data augmentation can truly help improving prediction performance.

The area under the receiver operating characteristic (ROC) curve, or AUC, has been used in medical diagnostics since the 1970s as an alternative single-number metric for measuring the predictive capacity of learning algorithms. It has been found to be a superior metric compared to accuracy, especially in data mining applications where ranking is important. AUC more effectively represents ranking than accuracy. [14] In the field of disease prediction, false negative diagnoses can have a severe consequence. It is important to take this into account when evaluating the results of imbalanced data learning in disease diagnosis. Therefore, using AUC as a metric can provide a more accurate assessment of the outcome

K-fold Cross Validation (CV) serves as a typical approach for performance estimation and model selection in the data mining and machine learning communities.[15] It was first proposed in the 1930s [16]. An explicit explanation of cross-validation, comparable to the present form of k-fold CV, emerged for the first time in 1968 by Mosteller and Turkey[17]. Stone et al. [18] used CV to choose appropriate model parameters, as opposed to just evaluating model performance. Nowadays, it is often used in applied machine learning to compare and choose a model for a specific predictive modeling issue because it is simple to comprehend, straightforward to implement, and yields skill estimates that are typically less biased than those from other approaches.[19]

K-fold Cross Validation (CV) is a common approach for evaluating the performance of machine learning models and selecting the best one for a specific predictive task. [15]It was first proposed in the 1930s [16]and has undergone several changes since then. The current form of k-fold CV was first explicitly described in 1968 by Mosteller and Turkey[17]. Stone et al. [18] used CV to choose appropriate model parameters, as opposed to just evaluating model performance. In k-fold cross-validation, the data is first divided into k segments or "folds" of equal size. The process is then repeated k times, with a different fold being used as the validation set each time, while the remaining k-1 folds are used for training. Stratified k-fold cross-validation is a variation of this method that ensures that each fold has a similar distribution of the output variable by using a stratified sampling approach. This helps to ensure that the validation set is representative of the overall population. The technique is often used in applied machine learning due to its simplicity and ability to yield less biased estimates of model performance.[20]

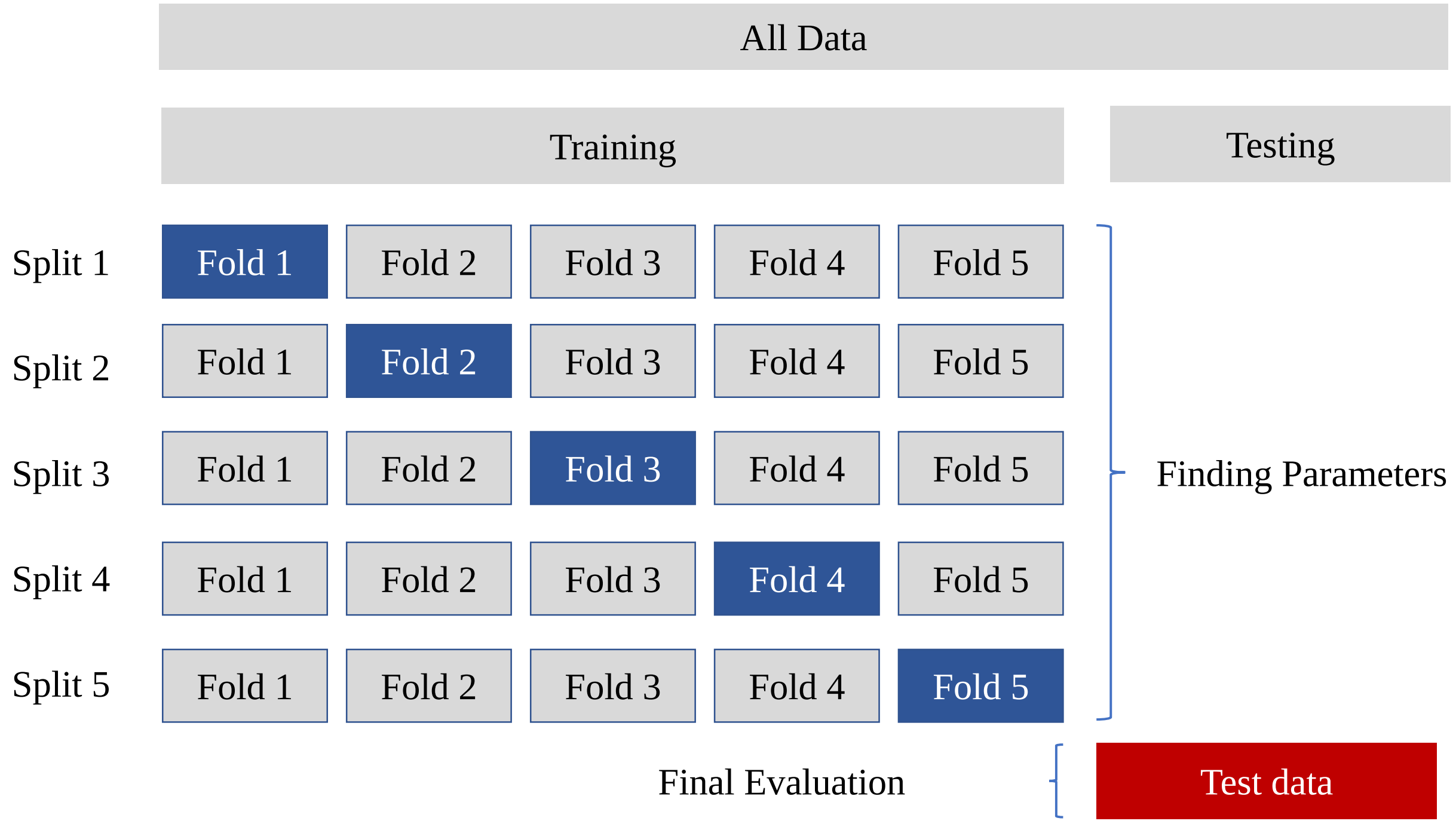

Figure 1 Traditional K-fold Cross Validation Visualization[21]

Previous papers tried to compare the performance of models learned from augmented data and the ones learned from unrevised data using k-fold cross validation **[5].** The imbalanced-learn package documentation Plotting the Validation Curves[22](access on August 8th 2022) gives an example that puts SMOTE augmentation method and DecisionTreeClassifier in a pipeline and then calculated cross-validation test score based on augmented data; the package provides an evaluation tutorial as well, which combines SMOTE and linear SVM model into a pipeline, fits the model, and evaluates prediction results. [23] Here we call the evaluation used by the sklearn imbalanced-learn package a Traditional evaluation method. Since the augmented dataset contains synthetic data points that were added to the original dataset to make it balanced, an issue is raised as to how the synthetic data points should be excluded from participating in the "testing" procedure. The synthetic data point" is "manufactured" based on the original data and will very likely favor a model that is learned from the same data, a phenomenon that is so-called the "cyclic effect." Due to this, the synthetic data included in the augmented dataset should not be used in testing a model during the evaluation process. This is noted by researchers, online discussions, and books[24-26]. When applying the k-fold CV to the augmented dataset, all the synthetic data will have an opportunity to test a model because each of the k-folds will be used as the testing data once. This may significantly but deceivingly increase the mean testing AUC, which has been widely used as an indicator of the prediction performance of the output model. Fotouhi et al. noticed this issue and kept testing the dataset before augmenting the dataset when doing the imbalanced data learning evaluation. [27] However, this evaluation did not clarify if they kept the same imbalanced rates in testing data and cannot make sure all the data have equal opportunity to be tested[27]. Therefore, it is essential to design a unique k-fold CV scheme suitable for evaluating models learned from an augmented dataset. Based on our literature search so far, we have not found an existing k-fold CV scheme that does this.

Using CV to evaluate imbalanced dataset learning does not have a correct standard pipeline now. The imbalanced-learn package documentation provides an example of this approach by combining the SMOTE augmentation method and DecisionTreeClassifier in a pipeline, then calculating the cross-validation test score based on the augmented data. This approach, known as the traditional evaluation method, raises concerns about how to exclude synthetic data points from the testing procedure. These synthetic data points are generated based on the original data and may model performance in favor of the augmented data. To address this, researchers have proposed excluding synthetic data from the testing process and testing the dataset before augmenting it. However, these evaluations do not ensure that the same imbalanced rates are maintained in the testing data. Thus, there is a need to develop a unique k-fold cross-validation scheme that is suitable for evaluating models trained on augmented data. So far, no such method has been found in the existing literature.

In this research, we propose a new k-fold CV framework that excludes synthetic data from testing a model. We implemented the new framework in python and tested it with 6 data balancing methods, 11 existing machine learning methods, and 8 different datasets. To demonstrate the model performance differences using the new and old evaluation frameworks and control for other factors, we used default parameters for each machine learning method in sklearn and package imbalanced-learn.

# METHODS

## Data Balancing Methods

The primarily widely used approaches for balancing data are oversampling (data augmentation) and under-sampling. In this study, we included 6 existing data balancing methods to test our evaluation framework, including the ROS, RUS, SMOTE, SVMSMOTE, ADYSYN, and the Cluster Centroids method. A description including the parameters and references of each of the six methods is listed in Appendix Table 1.

When dealing with datasets that are wildly out of balance, resampling is a common solution. Samples from the majority class are discarded (under-sampling), and/or minority class instances are increased (oversampling). SMOTE, SVMSMOTE, ADASYN, ROS, RUS, and ClusterCentroids are data resampling methods implemented in the sklearn imblearn package [28] These tactics, notwithstanding their benefits in terms of class balance, have several downsides (there is no free lunch). Overfitting may occur due to oversampling, which is as simple as duplicating random records from the minority class. Remove random records from the majority class to reduce under-sampling, which may result in data loss.

The data balancing methods we used in our new evaluation framework include the following:
**Random Over sample:** A method that increases the number of instances in the minority class by randomly replicating them in order to present a higher representation of the minority class in the sample.
**Random Under sampler:** A method that aims to balance class distribution by randomly eliminating majority class examples. This is done until the majority and minority class instances are balanced.
**SMOTE:** In a traditional oversampling strategy, minority data are reproduced from the population of minority data. While this enhances the data set, it adds no additional information or variety to the machine learning model. SMOTE generates synthetic data using a k-nearest neighbor method. SMOTE begins with selecting random data from the minority class and then sets k-nearest neighbors. The random data would then be combined with the randomly chosen k-nearest neighbors to create synthetic data
**SVMSMOTE:** SVMSMOTE is a variation of the SMOTE method that uses an SVM algorithm to identify samples to be employed in generating new synthetic samples.
**ADASYN:** ADASYN is another variation from SMOTE; ADASYN creates synthetic data according to the data density. The synthetic data generation would be inversely proportional to the density of the minority class. It means more synthetic data are created in regions of the feature space where the density of minority examples is low and fewer or none where the density is high.
**Cluster Centroids:** Under sample cluster centroids by creating centroids using clustering algorithms and under-sampling the majority class by substituting the cluster centroid of a K-Means algorithm for a cluster of majority samples. This approach maintains N majority samples by fitting the K-Means algorithm with N clusters to the majority class and then utilizing the coordinates of the N cluster centroids as fresh majority samples.

**Machine Learning methods**

We included 11 different existing machine methods in this study. We conjecture that the prediction performance of a machine-learning method is dataset dependent, and some methods might be more sensitive to data imbalance and data augmentation than other methods. We aimed to see how our new evaluation framework influences each of the machine methods in terms of the mean testing AUC. We listed the hyperparameters we used for each method below in Appendix Table 2 Parameters of Machine Learning Methods. All machine learning models used the sklearn package and seed =20211228; using the same parameters can help with replicating this study.

**Logistic regression:** Logistic regression is a classification algorithm for supervised learning that predicts the likelihood of a target variable. Logistic regression is commonly employed when target or dependent variable characteristics are dichotomous. [29]

**Nearest Neighbors:** Neighbors-based classification is a kind of instance-based learning or non-generalizing learning: it does not seek to build a general internal model but instead saves instances of the training data. Classification is determined by a simple majority vote of closest neighbors for each point: a query point is allocated the data class with the most representation among its nearest neighbors. k we used here is 5. [30]

**Linear Support Vector Machine:** Support vector machines (SVMs) are a set of machine learning algorithms that were first developed for the classification issue and then adapted for use in other contexts. The most fundamental benefit of an SVC is with a linear kernel, in which the decision boundary is a straight line or hyperplane in higher dimensions. Linear kernels are seldom used in reality, but I wanted to demonstrate it here since it is the simplest form of SVC. As seen below, it is not very good at categorizing since the data is not linear.[31]

**Radial Basis Function kernel Support Vector Machine:** RBFSVM's decision boundary is nonlinear. We use the default kernel "rbf."

Radial Basis Function $k(x, x') = \exp\left(\frac{-||x-x'||\hat{}2}{2\sigma\hat{}2}\right)$ where $||x-x'||\hat{}2$ is the squared Euclidean distance between two data points $x$ and $x'$.[31]

**Decision Tree:** Decision Trees (DTs) are a non-parametric approach of supervised learning used for classification and regression. The objective is to develop a model capable of predicting the value of a target variable using simple decision rules learned from the data attributes.[32,33]

**Neural Net:** Multilayer Perceptron (MLP)is a feedforward artificial neural network model that translates input data sets to an appropriate set of outputs. An MLP comprises numerous layers, each of which is completely coupled to the subsequent layer. Class MLPClassifier provides an MLP algorithm that is trained via Backpropagation.[34]

**AdaBoostClassifier:** An AdaBoost classifier is a meta-estimator that begins by fitting a classifier to the original dataset, followed by fitting additional copies of the classifier to the same dataset, with the weights of incorrectly classified instances adjusted such that subsequent classifiers concentrate more on challenging cases.[35,36]

**Random Forest Classifier:** "Random forest" is an ensemble machine learning method that utilizes multiple decision trees to create a more robust and accurate model. It works by training numerous decision trees on subsets of the data, known as bootstrap samples, and combining their predictions through averaging or majority voting. The number of decision trees is controlled by the n_estimators parameter, and the size of the bootstrap samples is controlled by the max_samples parameter[37]. The method was first introduced by Breiman in 2001 [38] and has since become a popular choice for both classification and regression tasks.

**Gradient Boosting:** Gradient Boosting classifier constructs an additive model in a stage-by-stage method; it permits the optimization of any loss functions that are differentiable. Based on the negative gradient of the binomial or multinomial deviance loss function, n-class regression trees are fitted at each step. In the case of binary classification, just one regression tree is generated.**[39]**

**Naïve Bayes:** Naïve Bayes techniques are a collection of supervised learning algorithms based on Bayes theorem with the "naive" assumption of conditional independence between every pair of features given the class variable value. The primary difference between the various Naïve Bayes classifiers is the assumptions they make on the distribution. GaussianNB implements the classification method Gaussian Naïve Bayes. It is assumed that the probability of the characteristics is Gaussian.[40]

**Quadratic Discriminant Analysis:** Quadratic Discriminant Analysis is a classifier with a quadratic decision boundary that is produced by fitting class conditional densities to the data and Bayes'Bayes' rule. Each class is assigned a Gaussian density by the model.**[41]**

## The Evaluation Framework for Learning from Augmented Data (EFLAD) concerning k-fold CV

We propose a 5-fold cross-validation evaluation framework for evaluating the prediction results of imbalanced data learning. The framework visualization is presented in Figure 2. Illustration of EFLAD, the Evaluation Framework for Learning from Augmented Data., Illustration of EFIDL. The proposed framework is divided into four steps: First, split the imbalanced dataset into five equal portions using a stratified strategy, ensuring that each portion has the same imbalanced rate as the whole dataset. This ensures that each portion represents the attributes of the entire dataset. Second, apply data augmentation methods on four portions of the data to obtain the augmented dataset. Third, train machine learning models using the augmented dataset. Fourth, use the remaining portion of the original data before augmentation in model evaluation to obtain the AUC for that portion. Repeat this process for all five data portions, and finally, take the mean of the AUCs as the final AUC. For example, we can split the data into P1, P2, P3, P4, and P5. We first keep P2, P3, P4, and P5 in the training dataset, apply data augmentation methods on the training dataset, and then use P1 to test the model performance and report the model AUC for P1. When using P2 in testing, we keep P1, P3, P4, and P5 in the training dataset, augment the training dataset, and then train ML models against the augmented training data.

After iterating the process five times, we take the mean of the testing AUCs and plot the AUC-ROC to analyze the effect of data augmentation.

In this evaluation framework, the original dataset is split into subsets that maintain the original imbalanced rate. The machine learning models are trained on a balanced dataset after augmentation, and each portion of the original data has an equal opportunity to be tested in the evaluation. The final AUC result represents the robustness of the evaluation.

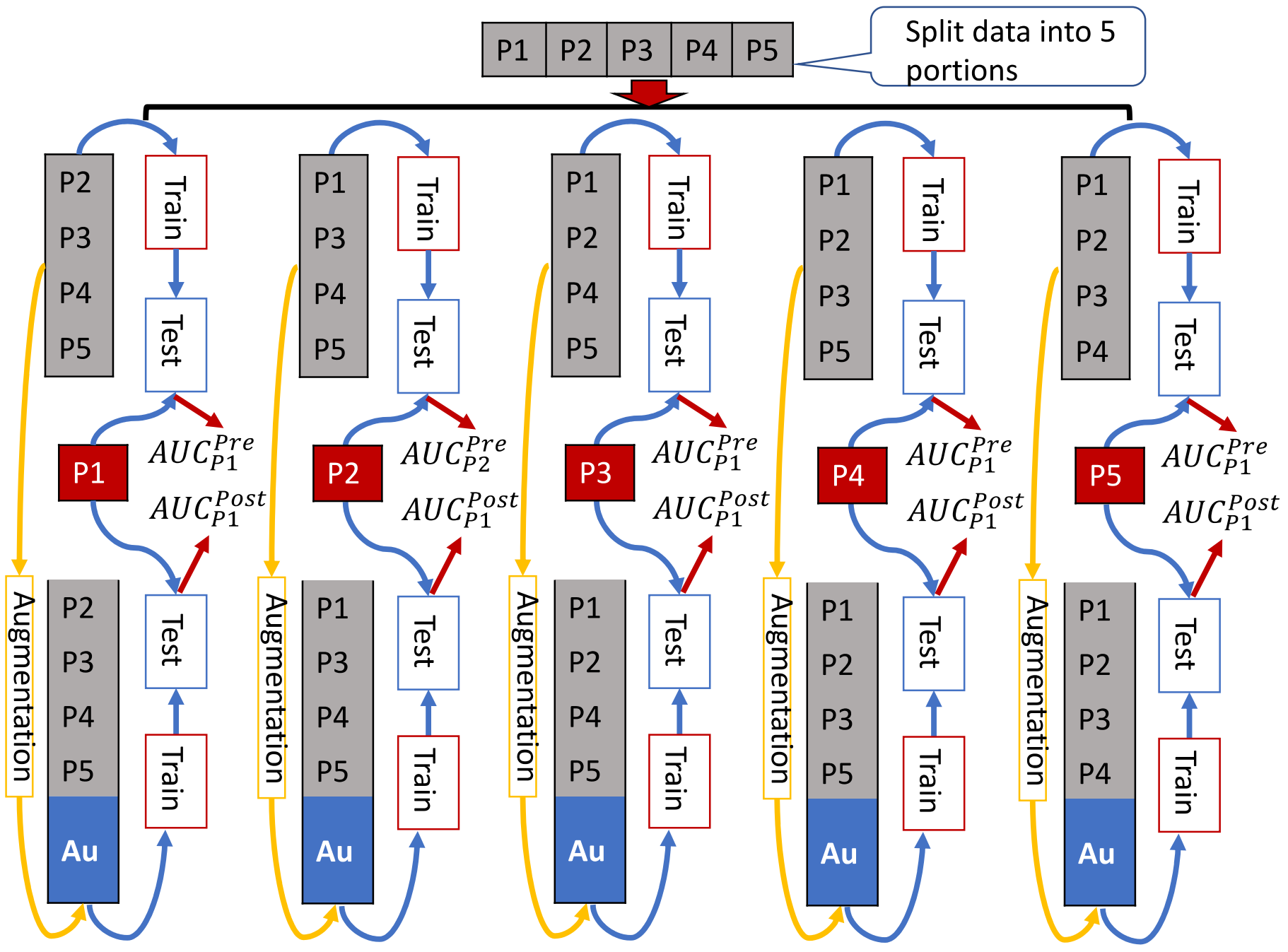

**Figure 2.** Illustration of EFLAD, the Evaluation Framework for Learning from Augmented Data

## Datasets

We included eight datasets in the EFLAD evaluation. All datasets are health-related; dataset links and original papers can be found in the dataset description table.

**LSM-5Year:** The Lynn Sage Metastasis (LSM) datasets were developed and published via previous studies[42,43]. The LSM-5Year dataset contains a binary outcome variable that indicates whether the patient metastasizes within five years. It includes 31 candidate predictors, including race, ethnicity, smoking, alcohol_useage, family_history, age_at_diagnosis, menopause_status, TNEG, ER, ER_percent, PR, PR_percent, P53, HER2, HER2_percent, Ki67, Ki67_percent, oncotype_score, t_tnm_stage, n_tnm_stage, stage, lymph_node_removed, lymph_node_ positive, lymph_node_status, histology, size, grade, invasive, DCIS_level, re_excision, and surgical_margins. This dataset has 437 cases and 3752 controls for a total of 4189 records.

**Wisconsin-Original:** The Wisconsin Breast Cancer -Original dataset was collected by Dr.William H. Wolberg using the fine needle aspiration (FNA) technic. There are 699 instances in the dataset, divided into two classes (malignant and benign) and nine integer-valued attributes. 16 cases from the dataset have missing values. The original paper did not report the same data number as the dataset. This dataset does not have demographic information.[44]

**Wisconsin-Diagnosis:** This dataset is published on the UCI Machine Learning data repository. It contains 569 samples that were collected from the FNA technic; all malignant aspirates were confirmed, whereas FNAs diagnosed as benign masses were biopsied only patient'stient's request.
The paper demonstrated that they deleted 18 benign cases that have identical characteristics with malignant cases in limitation. [45]

**SEER:** To lower the cancer burden among the U.S. population, the Surveillance, Epidemiology, and End Results (SEER) Program offers data about cancer statistics. SEER is financed by the Division of Cancer Control and PopuSciences'iences' Surveillance Research Program. [46] A researcher at Jilin University published this breast cancer subset of SEER data on the IEEE data portal. The data were extracted from the SEER program 2017 November update; SEER primary cites recode NOS histology codes 8522/3, diagnosed in 2006-2010.[46] 4024 patients with 15 features were ultimately included in this dataset.

**Segment-PCP_4:** There are six segment-PCP data sets that were constructed using the AAindex database and published protein sequence data.[47,48] Segment-PCP_4 is 4608 27-AA protein sequence segments derived from all K sites in 203 proteins (with sequences saved in FASTA format) comprising empirically confirmed ubiquitination sites utilized in a paper by Chen et al.[49]

**Segment-PCP_5:** In Data Set Segment-PCP_5, authors extracted all K sites from 96 protein sequences (in FASTA format) collected from experiments and published literature by the authors in[50], resulting in 3651 27-AA segments with 131 ubiquitination-positive segments and 3520 non-ubiquitination-positive segments.

**Segment-PCP_6:** Segment-PCP_6 is a collection of 676 27-AA protein sequence segments (with 37 ubiquitination ones and 639 non-ubiquitination ones, and all with central K sites), extracted from 21 protein sequences (in FASTA format) containing purportedly verified ubiquitination sites that were used as independent testing data in[49]

**Pima-Diabetes:** The Diabetes in Pima Indians dataset was compiled by the National Institute of Diabetes and Digestive and Kidney Diseases. The purpose of the dataset is to predict whether or not a patient has diabetes based on certain diagnostic metrics included within the information. The selection of these examples from a broader database was subject to a number of restrictions. All female patients here are at least 21 years old and of Pima Indian descent. There are just 268 positive class samples. It will be ideal for the classifier to have a high sensitivity for detecting diabetes cases.[51]

Table 1 Datasets Description

| Data Set | #Cases | #Features | Outcome/Class | Imbalanced-Rate* |
|---|---|---|---|---|
| LSM-5 Year [42,43] | 4189 | 31 | Distance Recurrence<br>3752(89.6%) recurrence<br>437 (10.4%) non-Recurrence | 0.16 |
| Wisconsin-Original[44] | 699 | 9 | tumor biopsy result<br>458 (65.5%) Benign<br>241(34.5%) Malignant | 0.52 |
| Wisconsin-Diagnostic[52] | 569 | 30 | tumor biopsy result<br>357 (62.7%) Benign,<br>212(37.3%) Malignant | 0.59 |
| SEER- IEEE portal[46] | 4024 | 15 | Survival status<br>Dead : 616(15.3%)<br>Alive:3405 84.7%) | 0.18 |
| SegmentPCP_4[53] | 4608 | 530 | containing<br>263 ubiquitination positive segments and<br>4345 non-ubiquitination ones | 0.06 |
| SegmentPCP_5[53] | 3651 | 530 | 131 ubiquitination positive segments and<br>3520 non-ubiquitination negative ones | 0.04 |
| SegmentPCP_6[53] | 676 | 530 | 37 ubiquitination positive segments and<br>639 non-ubiquitination negative ones | 0.06 |
| Diabetes[54](Pima) | 768 | 8 | 268 are class 1<br>500 are class 0 | 0.536 |

* Imbalanced-Rate(IR) is the number of minorities divided by the number of majority cases. If a dataset is balanced, the imbalanced rate is 1; the small IR is, the more imbalanced the dataset is.

# RESULTS

Table 3 below shows a comparison of results between using the 5-fold CV with the EFLAD framework and the traditional 5-fold CV framework for different types of augmented data across all machine learning methods, which were used to learn prediction models from the LSM-5year dataset. Comparison of TRA and EFLDA for other datasets results, see appendix Tables

<See all datasets result in summary in the Appendix>

| LSM -5 Years | | | | | | | | | | | | | |
|---|---|---|---|---|---|---|---|---|---|---|---|---|---|
| | | **LR** | **KNN** | **SVM_L** | **SVM_R** | **DT** | **RF** | **NN** | **ADB** | **GB** | **NB** | **QDA** | **Avg** |
| **BEF** | **EFLDA** | 0.759 | 0.658 | 0.580 | 0.655 | 0.710 | 0.776 | 0.695 | 0.746 | 0.835 | 0.748 | 0.737 | 0.718 |
| | **TRA** | 0.759 | 0.658 | 0.580 | 0.655 | 0.710 | 0.776 | 0.695 | 0.746 | 0.835 | 0.748 | 0.737 | 0.718 |
| **ADASYN** | **EFLDA** | 0.643 | 0.667 | 0.642 | 0.647 | 0.696 | 0.714 | 0.641 | 0.645 | 0.835 | 0.615 | 0.605 | 0.668 |
| | **TRA** | 0.813 | 0.895 | 0.811 | 0.903 | 0.733 | 0.823 | 0.877 | 0.706 | 0.765 | 0.780 | 0.810 | 0.811 |
| | **% diff (EFLDA)** | -15.261 | 1.464 | 10.703 | -1.253 | -1.977 | -8.015 | -7.757 | -13.536 | -0.029 | -17.862 | -17.896 | -6.493 |
| | **% diff (TRA)** | 7.121 | 36.113 | 39.835 | 37.871 | 3.193 | 6.033 | 26.153 | -5.285 | -8.333 | 4.254 | 9.946 | 14.264 |
| **SMOTE** | **EFLDA** | 0.650 | 0.674 | 0.647 | 0.649 | 0.698 | 0.721 | 0.646 | 0.682 | 0.835 | 0.627 | 0.609 | 0.676 |
| | **TRA** | 0.829 | 0.930 | 0.827 | 0.966 | 0.750 | 0.840 | 0.906 | 0.715 | 0.776 | 0.791 | 0.839 | 0.834 |
| | **% diff (EFLDA)** | -14.385 | 2.496 | 11.620 | -0.909 | -1.794 | -7.112 | -7.094 | -8.515 | -0.035 | -16.170 | -17.260 | -5.378 |
| | **% diff (TRA)** | 9.235 | 41.370 | 42.628 | 47.377 | 5.636 | 8.292 | 30.269 | -4.127 | -7.079 | 5.786 | 13.858 | 17.568 |
| **SVMSMOTE** | **EFLDA** | 0.710 | 0.673 | 0.708 | 0.661 | 0.658 | 0.757 | 0.665 | 0.739 | 0.816 | 0.688 | 0.616 | 0.699 |
| | **TRA** | 0.894 | 0.946 | 0.893 | 0.967 | 0.834 | 0.899 | 0.931 | 0.849 | 0.882 | 0.862 | 0.887 | 0.895 |
| | **% diff (EFLDA)** | -6.427 | 2.266 | 22.101 | 0.874 | -7.299 | -2.448 | -4.427 | -0.951 | -2.323 | -8.053 | -16.338 | -2.093 |
| | **% diff (TRA)** | 17.847 | 43.901 | 53.944 | 47.521 | 17.461 | 15.876 | 33.910 | 13.828 | 5.669 | 15.215 | 20.477 | 25.968 |
| **ROS** | **EFLDA** | 0.758 | 0.649 | 0.762 | 0.653 | 0.714 | 0.773 | 0.671 | 0.733 | 0.835 | 0.748 | 0.733 | 0.730 |
| | **TRA** | 0.776 | 0.946 | 0.771 | 1.000 | 0.733 | 0.819 | 0.910 | 0.732 | 0.835 | 0.752 | 0.797 | 0.825 |
| | **% diff (EFLDA)** | -0.183 | -1.241 | 31.319 | -0.330 | 0.476 | -0.330 | -3.503 | -1.720 | -0.021 | -0.024 | -0.526 | 2.174 |
| | **% diff (TRA)** | 2.194 | 43.900 | 32.874 | 52.600 | 3.146 | 5.551 | 30.879 | -1.861 | 0.020 | 0.576 | 8.200 | 16.189 |
| **RUS** | **EFLDA** | 0.750 | 0.714 | 0.754 | 0.498 | 0.707 | 0.761 | 0.659 | 0.738 | 0.835 | 0.743 | 0.731 | 0.717 |
| | **TRA** | 0.725 | 0.724 | 0.728 | 0.498 | 0.669 | 0.747 | 0.669 | 0.728 | 0.832 | 0.731 | 0.721 | 0.706 |
| | **% diff (EFLDA)** | -1.140 | 8.613 | 30.021 | -24.054 | -0.457 | -1.861 | -5.211 | -1.034 | -0.024 | -0.695 | -0.753 | 0.309 |
| | **% diff (TRA)** | -4.466 | 10.028 | 25.496 | -24.038 | -5.862 | -3.679 | -3.857 | -2.403 | -0.355 | -2.249 | -2.168 | -1.232 |
| **CC** | **EFLDA** | 0.655 | 0.679 | 0.659 | 0.502 | 0.626 | 0.673 | 0.641 | 0.660 | 0.835 | 0.634 | 0.620 | 0.653 |
| | **TRA** | 0.958 | 0.939 | 0.955 | 0.505 | 0.901 | 0.943 | 0.956 | 0.890 | 0.893 | 0.940 | 0.946 | 0.893 |
| | **% diff (EFLDA)** | -13.701 | 3.302 | 13.606 | -23.442 | -11.814 | -13.229 | -7.780 | -11.526 | -0.058 | -15.193 | -15.786 | -8.693 |
| | **% diff (TRA)** | 26.198 | 42.824 | 64.730 | -22.986 | 26.901 | 21.554 | 37.429 | 19.347 | 6.889 | 25.663 | 28.376 | 25.175 |

**Table 3:** BEF: before_resampling, ROS: Random Over Sampler, RUS: Rander Under Sampler, CC: Cluster Centroids, TRA: traditional 5-k cross-validation framework, EFLAD: an evaluation framework for learning from augmented data, LR: Logistic Regression, KNN: k-nearest Neighbor, $SVM_l$: Linear SVM, $SVM_r$: RBF SVM, DT: Decision Tree, RF: Random Forest, NN: Neural Network, ADB: AdaBoost, GB: Gradient Boosting, NB: Naïve Bayes, Avg: average

We found that data augmentation did not significantly improve the AUC in any of our eight datasets. The performance of the model largely depends on the characteristics of the dataset and the machine learning technique used. For example, in the LSM dataset, the best AUC was achieved before adopting any data augmentation techniques. Similarly, in the WBC original dataset, all AUC findings were more than 0.97, but we did not see a substantial improvement in the results by applying data augmentation techniques like SVMSMOTE or SMOTE. The best AUC for each dataset can be seen in the table below. For datasets such as Pima Diabetes, Adult Income, LSM, WBC_original, SegmentPCP 4, and SegmentPCP5, the best model was achieved using Gradient Boosting without data augmentation. For the SEER dataset, the best model was Linear SVM without data augmentation. The Neural Net after SVMSMOTE augmentation performed the best in the WBC_diagnostics data. And for SegmentPCP6, the best result was achieved using Gradient Boosting with the ADASYN augmentation method.

<Table 4 – Best performance algorithm for each dataset >

| Dataset | Best Model | Sampling method | Best AUC |
|---|---|---|---|
| Pima Diabetes | Gradient Boosting | Before resampling | 0.919717 |
| LSM | Gradient Boosting | Before resampling | 0.835072 |
| WBC_original | Gradient Boosting | Before resampling | 0.997566 |
| WBC_diagnostics | Neural Net | SVMSMOTE | 0.993531 |
| SEER | Linear SVM | Before resampling | 0.974785 |
| SegmentPCP4 | Gradient Boosting | Before resampling | 0.860274 |
| SegmentPCP5 | Gradient Boosting | Before resampling | 0.875686 |
| SegmentPCP6 | Gradient Boosting | ADASYN | 0.877462 |

We found that for the LSM 5-year dataset, none of the data augmentation methods helped with LR, RF, NN, ADB, GB, NB, and QDA algorithms. ADASYN improved KNN performance by 1.46%, SMOTE improved KNN by 2.50%, RUS improved KNN by 8.61%, and CC improved KNN by 3.30%. ADASYN helped SVM_L improve by 10%, SMOTE improved SVM_L by 11.62%, SVMSMOTE improved by 22.10%, ROS improved by 31.32%, and RUS improved by 30%. However, for the SEER dataset, none of the data augmentations helped with the SVM_L model. This performance depends on the dataset feature characteristics; the LSM and SEER datasets have similar imbalanced ratios and data points. However, SEER has 15 features, and LSM has 31 features.

We also compared our proposed evaluation framework (EFLAD) with the traditional evaluation method from the imbalanced-learn tutorial[23]. For the LSM-5-year dataset, we used the SMOTE resampling method and Logistic Regression. The traditional result was an F1 score of 0.78; however, when using EFLAD, the F1 score was 0.54. This comparison shows that adding augmented data into evaluation falsely boosts the model performance.

## DISCUSSION

In this study, we proposed an evaluation framework for imbalanced dataset learning. Our 5-fold stratified cross-validation framework aims to prevent the inclusion of synthetic data in the model evaluation process. This framework can be applied to evaluate the performance of machine learning algorithms not only in imbalanced data learning but also in any research involving synthetic data in model building, such as assessing missing data imputation algorithms. It is important for researchers to keep in mind that when building a prediction model using real-world healthcare data, the evaluation should also be conducted on the original dataset in order to maintain its characteristics.

Chawla et al. developed the SMOTE method for oversampling minority classes by creating synthetic instances, as opposed to oversampling with replacement. They generated synthetic examples in a more general way by working in "feature space" rather than "data space." [5] In SMOTE algorithms, minority classes are oversampled by inserting synthetic samples along the line segments connecting any or all of the k closest neighbors of the minority class. Since the introduction of SMOTE for dealing with imbalanced data, numerous SMOTE-based algorithms have been developed to improve imbalanced data learning, such as RCSMOTE, Borderline-SMOTE, and Safe-level-SMOTE. [55-57] As of August 2022, the SMOTE paper has received over 19,859 citations. The SMOTE approach has been shown to improve the accuracy of classifiers for minority classes in experimental results. Out of a total of 48 experiments performed, the SMOTE-classifier performed the best for 44 experiments. However, using our proposed evaluation framework, our experiments show that the AUC of the SMOTE-classifier approach does not significantly outperform the classifier-only (before data augmentation) approach.

Evaluating the outcomes of data augmentation should be approached with caution to prevent errors from being introduced. Some researchers may be uncertain about when and how to evaluate the results of data augmentation. We found an online tutorial that provided misleading code and inappropriate evaluation methods.[58]This example used synthesized data in the evaluation, resulting in a significant improvement in the prediction AUC after data augmentation on imbalanced learning. However, it is important to note that using synthesized data or balanced testing datasets in the evaluation of imbalanced learning can lead to false positive results. KSV discussed the differences between using "SMOTE in the pipeline" and "SMOTE out of the pipeline" [24,25]. Our proposed evaluation framework offers a robust evaluation method for assessing the performance of models trained on augmented data in imbalanced learning.

Our findings challenge the previous argument made by some researchers that data augmentation improves minority class learning results, including accuracy and AUC[59-61], as our results show that the model AUC is highly dependent on the data features and machine learning methods. In most experiments, data augmentation did not improve the AUC. In the Wisconsin breast cancer dataset, all machine learning methods performed well because the data has features that are highly associated with the outcome. The before-resampling logistic regression has an AUC of 0.995, which is already a perfect score for prediction. Nonetheless, for dataset Segment-PCP_6 with 530 features and 676 records, most machine learning methods could not learn a good model; 4 before resampling results outperform other data balancing methods AUC.
Our results reveal that previous researchers may have made incorrect conclusions based on improper evaluation methods.

Different combinations of data augmentation methods and machine learning classifiers respond differently depending on the characteristics of the dataset. For example, the Wisconsin Breast

Cancer dataset has been well studied in many research papers[7,62-64]. Studies have been found to benefit from data augmentation methods, with some improved reporting performance, even near perfection. However, in our experiment, we found that while SMOTE improved the ROC from 0.9954 to 0.9959, the dataset already had a strong signal for prediction, and the improvement was not clinically significant. On the other hand, for datasets that are more imbalanced, data augmentation did not show significant improvement in performance.

Our experiments showed that not all data augmentation methods could improve prediction model performance. On the WBC prognostics dataset, SMOTE improved performance for 8 out of 11 ML methods, whereas on the LSM dataset, it improved performance for only 2 out of 11 ML methods. ADASYN helped 1 out of 11 methods on the Pima-Diabetes dataset and 7 ML methods on the SegmentPCP4 data. SVMSMOTE helped improve 3 out of 11 ML methods on the LSM dataset and 7 out of 11 on the SegmentPCP4 data. ROS helped 8 out of 11 ML methods on the WBC Diagnostics and WBC prognostics datasets and 2 out of 11 ML methods on the LSM data. RUS helped 8 out of 11 ML methods on the SEER data and 2 on the LSM data. CC helped 7 out of 11 ML methods on the WBC original data and only 1 on the SEER data.

The experiment results show that the effectiveness of data augmentation methods varies depending on the dataset and machine learning methods used. In some datasets, such as the Wisconsin Breast Cancer prognostics dataset, data augmentation methods like SMOTE improved the performance of several machine learning methods. However, in other datasets like the LSM dataset, data augmentation methods did not show significant improvement in model performance. Additionally, different data augmentation methods had varying levels of effectiveness depending on the dataset, with some methods showing better results on specific datasets than others. Overall, the results suggest that data augmentation methods should be carefully evaluated and selected based on the particular characteristics of the dataset and the machine learning methods used.

For future studies, we will examine the mechanism of how data augmentation can affect prediction model performance. Our code is available on Google Colab[1] , and we encourage researchers to consider using our evaluation framework when building imbalanced data prediction models.

## LIMITATIONS:

We did not perform parameter selection for every machine learning method used in our experiments. This means that we did not optimize the parameters of each method to find the best-performing model for each dataset. Instead, we used the default parameters provided by the scikit-learn library. This could have affected the performance of the models, as different parameter settings may have led to better results. Additionally, each machine learning method performs differently based on the characteristics of the dataset, such as the number of features, the class imbalance ratio, and the amount of noise present in the data. For this reason, the before-sampling results for each machine learning method may not represent the best AUC that could be achieved with that method.

[1] https://drive.google.com/file/d/1P_LnxJOVTaTe7FB5a86lUd3yVyBXh0sy/view?usp=sharing

Furthermore, there are some results that show overfitting or AUC scores below 0.5. These results are included in our report, as they provide a comprehensive view of the performance of the different methods and data augmentation techniques. However, it is worth noting that these results may not represent the general performance of the models and may not be reliable for real-world applications.

**Acknowledgments**
This study was supported by the U.S. Department of Defense Award No. W81XWH1910495 (to X.J). Other than supplying funds, the funding agencies played no role in the research.

## ABBREVIATIONS

Area Under the Curve: AUC
Cross Validation: CV
Imbalanced-Rate: IR
Lynn Sage Dataset for Metastasis: LSM
Wisconsin Breast Cancer: WBC
The Surveillance, Epidemiology, and End Results Program: SEER

Random Under Sampling: RUS
Random Over Sampling: ROS
Synthetic Minority Over-sampling Technique: SMOTE
Adaptive Synthetic: ADASYN
ClusterCentroids: CC
Logistic Regression: LR
K-Nearest Neighbors: KNN
Linear Support Vector Machines: SVM_L
Radial Basis Function Kernel Support Vector Machines: SVM_R
Decision Tree: DT
Random Forest: RF
Neural Network  NN
Adaboost: ADB
Gradient Boost  GB
Naïve Bayesian:  NB
Quadratic Discriminant Analysis: QDA

# Appendix

## Table of Contents

# PARAMETERS

## Appendix Table 1 Parameters of Data Augmentation Methods

| Method name | Parameters |
|---|---|
| Random Over sampler | RandomOverSampler(random_state = seed) |
| Random Under Sampler | RandomUnderSampler(random_state = seed) |
| SMOTE[1] | SMOTE(random_state = seed) |
| SVMSMOTE[2] | SVMSMOTE(randome_states=seed) |
| ADASYN[3] | ADASYN(random_state = seed ) |
| Cluster Centroids | ClusterCentroids(random_state = seed) |

## Appendix Table 2 Parameters of Machine Learning Methods

| Method name | Parameters |
|---|---|
| Logistic Regression | LogisticRegression(random_state = seed) |
| Nearest Neighbors | KNeighborsClassifier(5) |
| Linear SVM | SVC(kernel="linear", C=0.025, random_state = seed,probability=True,) |
| RBF SVM | SVC(gamma=2, C=1,random_state = seed,probability=True,) |
| Decision Tree | DecisionTreeClassifier(max_depth=8, max_features='auto', max_leaf_nodes=15,min_samples_split=0.1, random_state=seed,splitter='random') |
| Random Forest | RandomForestClassifier(max_depth=5, n_estimators=10, random_state = seed) |
| Neural Net | MLPClassifier(hidden_layer_sizes=(20,), activation='relu', solver='adam', alpha=0.0001, max_iter=1500, random_state = seed) |
| AdaBoost | AdaBoostClassifier(n_estimators=50, learning_rate=0.01 , algorithm='SAMME.R', random_state= seed) |
| Gradient Boosting | GradientBoostingClassifier( loss='deviance', learning_rate=0.05, n_estimators=100, subsample=1.0,random_state=seed, max_features=None, verbose=0, max_leaf_nodes=None, warm_start=True, validation_fraction=0.1, n_iter_no_change=None, tol=0.0001, ccp_alpha=0.0), # Changed learing rate |
| Naïve Bayes | GaussianNB() |
| Quadratic Discriminant Analysis | QuadraticDiscriminantAnalysis() |

## Result Tables

## Appendix Table 3 LSM-5 Years Results

| LSM -5 Years | | | | | | | | | | | | | |
|---|---|---|---|---|---|---|---|---|---|---|---|---|---|
| | | **LR** | **KNN** | **SVM_L** | **SVM_R** | **DT** | **RF** | **NN** | **ADB** | **GB** | **NB** | **QDA** | **Avg** |
| **BEF** | **EFLDA** | 0.759 | 0.658 | 0.580 | 0.655 | 0.710 | 0.776 | 0.695 | 0.746 | 0.835 | 0.748 | 0.737 | 0.718 |
| | **TRA** | 0.759 | 0.658 | 0.580 | 0.655 | 0.710 | 0.776 | 0.695 | 0.746 | 0.835 | 0.748 | 0.737 | 0.718 |
| **ADASYN** | **EFLDA** | 0.643 | 0.667 | 0.642 | 0.647 | 0.696 | 0.714 | 0.641 | 0.645 | 0.835 | 0.615 | 0.605 | 0.668 |
| | **TRA** | 0.813 | 0.895 | 0.811 | 0.903 | 0.733 | 0.823 | 0.877 | 0.706 | 0.765 | 0.780 | 0.810 | 0.811 |
| | **% diff (EFLDA)** | -15.261 | 1.464 | 10.703 | -1.253 | -1.977 | -8.015 | -7.757 | -13.536 | -0.029 | -17.862 | -17.896 | -6.493 |
| | **% diff (TRA)** | 7.121 | 36.113 | 39.835 | 37.871 | 3.193 | 6.033 | 26.153 | -5.285 | -8.333 | 4.254 | 9.946 | 14.264 |
| **SMOTE** | **EFLDA** | 0.650 | 0.674 | 0.647 | 0.649 | 0.698 | 0.721 | 0.646 | 0.682 | 0.835 | 0.627 | 0.609 | 0.676 |
| | **TRA** | 0.829 | 0.930 | 0.827 | 0.966 | 0.750 | 0.840 | 0.906 | 0.715 | 0.776 | 0.791 | 0.839 | 0.834 |
| | **% diff (EFLDA)** | -14.385 | 2.496 | 11.620 | -0.909 | -1.794 | -7.112 | -7.094 | -8.515 | -0.035 | -16.170 | -17.260 | -5.378 |
| | **% diff (TRA)** | 9.235 | 41.370 | 42.628 | 47.377 | 5.636 | 8.292 | 30.269 | -4.127 | -7.079 | 5.786 | 13.858 | 17.568 |
| **SVMSMOTE** | **EFLDA** | 0.710 | 0.673 | 0.708 | 0.661 | 0.658 | 0.757 | 0.665 | 0.739 | 0.816 | 0.688 | 0.616 | 0.699 |
| | **TRA** | 0.894 | 0.946 | 0.893 | 0.967 | 0.834 | 0.899 | 0.931 | 0.849 | 0.882 | 0.862 | 0.887 | 0.895 |
| | **% diff (EFLDA)** | -6.427 | 2.266 | 22.101 | 0.874 | -7.299 | -2.448 | -4.427 | -0.951 | -2.323 | -8.053 | -16.338 | -2.093 |
| | **% diff (TRA)** | 17.847 | 43.901 | 53.944 | 47.521 | 17.461 | 15.876 | 33.910 | 13.828 | 5.669 | 15.215 | 20.477 | 25.968 |
| **ROS** | **EFLDA** | 0.758 | 0.649 | 0.762 | 0.653 | 0.714 | 0.773 | 0.671 | 0.733 | 0.835 | 0.748 | 0.733 | 0.730 |
| | **TRA** | 0.776 | 0.946 | 0.771 | 1.000 | 0.733 | 0.819 | 0.910 | 0.732 | 0.835 | 0.752 | 0.797 | 0.825 |
| | **% diff (EFLDA)** | -0.183 | -1.241 | 31.319 | -0.330 | 0.476 | -0.330 | -3.503 | -1.720 | -0.021 | -0.024 | -0.526 | 2.174 |
| | **% diff (TRA)** | 2.194 | 43.900 | 32.874 | 52.600 | 3.146 | 5.551 | 30.879 | -1.861 | 0.020 | 0.576 | 8.200 | 16.189 |
| **RUS** | **EFLDA** | 0.750 | 0.714 | 0.754 | 0.498 | 0.707 | 0.761 | 0.659 | 0.738 | 0.835 | 0.743 | 0.731 | 0.717 |
| | **TRA** | 0.725 | 0.724 | 0.728 | 0.498 | 0.669 | 0.747 | 0.669 | 0.728 | 0.832 | 0.731 | 0.721 | 0.706 |
| | **% diff (EFLDA)** | -1.140 | 8.613 | 30.021 | -24.054 | -0.457 | -1.861 | -5.211 | -1.034 | -0.024 | -0.695 | -0.753 | 0.309 |
| | **% diff (TRA)** | -4.466 | 10.028 | 25.496 | -24.038 | -5.862 | -3.679 | -3.857 | -2.403 | -0.355 | -2.249 | -2.168 | -1.232 |
| **CC** | **EFLDA** | 0.655 | 0.679 | 0.659 | 0.502 | 0.626 | 0.673 | 0.641 | 0.660 | 0.835 | 0.634 | 0.620 | 0.653 |
| | **TRA** | 0.958 | 0.939 | 0.955 | 0.505 | 0.901 | 0.943 | 0.956 | 0.890 | 0.893 | 0.940 | 0.946 | 0.893 |
| | **% diff (EFLDA)** | -13.701 | 3.302 | 13.606 | -23.442 | -11.814 | -13.229 | -7.780 | -11.526 | -0.058 | -15.193 | -15.786 | -8.693 |
| | **% diff (TRA)** | 26.198 | 42.824 | 64.730 | -22.986 | 26.901 | 21.554 | 37.429 | 19.347 | 6.889 | 25.663 | 28.376 | 25.175 |

## Appendix Table 4 WBC Original Results

| **WBC Original** | | | | | | | | | | | | | |
|---|---|---|---|---|---|---|---|---|---|---|---|---|---|
| | | **LR** | **KNN** | **SVM_L** | **SVM_R** | **DT** | **RF** | **NN** | **ADB** | **GB** | **NB** | **QDA** | **Avg** |
| **BEF** | **EFLDA** | 0.995 | 0.982 | 0.995 | 0.982 | 0.981 | 0.990 | 0.995 | 0.984 | 0.998 | 0.986 | 0.983 | 0.988 |
| | **TRA** | 0.995 | 0.982 | 0.995 | 0.982 | 0.981 | 0.990 | 0.995 | 0.984 | 0.998 | 0.986 | 0.983 | 0.988 |
| **ADASYN** | **EFLDA** | 0.995 | 0.983 | 0.995 | 0.980 | 0.977 | 0.990 | 0.992 | 0.974 | 0.997 | 0.985 | 0.982 | 0.986 |
| | **TRA** | 0.991 | 0.985 | 0.991 | 0.984 | 0.972 | 0.990 | 0.990 | 0.974 | 0.996 | 0.979 | 0.972 | 0.984 |
| | **% diff (EFLDA)** | 0.000 | 0.078 | -0.009 | -0.114 | -0.421 | 0.064 | -0.278 | -1.033 | -0.038 | -0.104 | -0.036 | -0.172 |
| | **% diff (TRA)** | -0.397 | 0.267 | -0.390 | 0.217 | -0.902 | 0.021 | -0.477 | -1.029 | -0.204 | -0.703 | -1.083 | -0.425 |
| **SMOTE** | **EFLDA** | 0.996 | 0.980 | 0.995 | 0.981 | 0.982 | 0.989 | 0.995 | 0.981 | 0.993 | 0.986 | 0.983 | 0.987 |
| | **TRA** | 0.996 | 0.990 | 0.996 | 0.993 | 0.982 | 0.993 | 0.996 | 0.982 | 0.984 | 0.986 | 0.983 | 0.989 |
| | **% diff (EFLDA)** | 0.019 | -0.174 | 0.009 | -0.038 | 0.074 | -0.095 | -0.024 | -0.308 | -0.413 | 0.012 | 0.007 | -0.085 |
| | **% diff (TRA)** | 0.037 | 0.867 | 0.019 | 1.125 | 0.110 | 0.333 | 0.054 | -0.148 | -1.346 | 0.027 | 0.012 | 0.099 |
| **SVMSMOTE** | **EFLDA** | 0.996 | 0.984 | 0.996 | 0.980 | 0.984 | 0.993 | 0.992 | 0.976 | 0.997 | 0.985 | 0.982 | 0.988 |
| | **TRA** | 0.993 | 0.985 | 0.993 | 0.987 | 0.976 | 0.989 | 0.994 | 0.975 | 0.993 | 0.980 | 0.975 | 0.985 |
| | **% diff (EFLDA)** | 0.057 | 0.170 | 0.019 | -0.129 | 0.372 | 0.381 | -0.296 | -0.834 | -0.009 | -0.085 | -0.031 | -0.035 |
| | **% diff (TRA)** | -0.221 | 0.354 | -0.279 | 0.549 | -0.513 | -0.046 | -0.166 | -0.873 | -0.438 | -0.623 | -0.797 | -0.278 |
| **ROS** | **EFLDA** | 0.996 | 0.983 | 0.995 | 0.982 | 0.981 | 0.991 | 0.995 | 0.982 | 0.993 | 0.986 | 0.983 | 0.988 |
| | **TRA** | 0.996 | 0.990 | 0.996 | 0.993 | 0.983 | 0.993 | 0.996 | 0.983 | 0.995 | 0.986 | 0.983 | 0.990 |
| | **% diff (EFLDA)** | 0.028 | 0.146 | 0.000 | 0.053 | 0.055 | 0.149 | -0.009 | -0.204 | -0.413 | 0.002 | 0.000 | -0.018 |
| | **% diff (TRA)** | 0.080 | 0.771 | 0.062 | 1.169 | 0.214 | 0.355 | 0.094 | -0.111 | -0.220 | 0.046 | 0.057 | 0.229 |
| **RUS** | **EFLDA** | 0.996 | 0.988 | 0.995 | 0.981 | 0.981 | 0.992 | 0.996 | 0.982 | 0.991 | 0.985 | 0.982 | 0.988 |
| | **TRA** | 0.993 | 0.983 | 0.992 | 0.974 | 0.973 | 0.989 | 0.992 | 0.969 | 0.992 | 0.983 | 0.979 | 0.984 |
| | **% diff (EFLDA)** | 0.033 | 0.570 | 0.000 | -0.043 | -0.027 | 0.206 | 0.028 | -0.211 | -0.643 | -0.024 | -0.109 | -0.020 |
| | **% diff (TRA)** | -0.290 | 0.102 | -0.294 | -0.751 | -0.787 | -0.116 | -0.325 | -1.508 | -0.566 | -0.254 | -0.334 | -0.466 |
| **CC** | **EFLDA** | 0.996 | 0.984 | 0.996 | 0.982 | 0.977 | 0.991 | 0.995 | 0.982 | 0.991 | 0.991 | 0.989 | 0.989 |
| | **TRA** | 0.986 | 0.972 | 0.986 | 0.966 | 0.964 | 0.976 | 0.986 | 0.957 | 0.980 | 0.971 | 0.968 | 0.974 |
| | **% diff (EFLDA)** | 0.009 | 0.229 | 0.014 | 0.033 | -0.361 | 0.156 | -0.033 | -0.169 | -0.657 | 0.524 | 0.635 | 0.035 |
| | **% diff (TRA)** | -0.943 | -0.992 | -0.974 | -1.572 | -1.734 | -1.378 | -0.881 | -2.716 | -1.754 | -1.473 | -1.523 | -1.449 |

## Appendix Table 5 WBC Diagnostics Results

| **WBC diagnostics** | | | | | | | | | | | | | |
|---|---|---|---|---|---|---|---|---|---|---|---|---|---|
| | | **LR** | **KNN** | **SVM_L** | **SVM_R** | **DT** | **RF** | **NN** | **ADB** | **GB** | **NB** | **QDA** | **Avg** |
| **BEF** | **EFLDA** | 0.989 | 0.966 | 0.983 | 0.964 | 0.932 | 0.978 | 0.992 | 0.956 | 0.993 | 0.973 | 0.979 | 0.973 |
| | **TRA** | 0.989 | 0.966 | 0.983 | 0.964 | 0.932 | 0.978 | 0.992 | 0.956 | 0.993 | 0.973 | 0.979 | 0.973 |
| **ADASYN** | **EFLDA** | 0.990 | 0.959 | 0.983 | 0.955 | 0.945 | 0.976 | 0.993 | 0.944 | 0.990 | 0.975 | 0.979 | 0.972 |
| | **TRA** | 0.972 | 0.968 | 0.952 | 0.977 | 0.910 | 0.972 | 0.989 | 0.890 | 0.983 | 0.935 | 0.963 | 0.955 |
| | **% diff (EFLDA)** | 0.099 | -0.740 | -0.035 | -0.905 | 1.312 | -0.184 | 0.046 | -1.166 | -0.367 | 0.183 | -0.032 | -0.163 |
| | **% diff (TRA)** | -1.701 | 0.214 | -3.114 | 1.340 | -2.432 | -0.643 | -0.316 | -6.888 | -1.048 | -3.964 | -1.659 | -1.837 |
| **SMOTE** | **EFLDA** | 0.990 | 0.965 | 0.983 | 0.965 | 0.946 | 0.976 | 0.991 | 0.961 | 0.986 | 0.974 | 0.976 | 0.974 |
| | **TRA** | 0.991 | 0.981 | 0.985 | 0.977 | 0.948 | 0.987 | 0.995 | 0.963 | 0.988 | 0.974 | 0.985 | 0.979 |
| | **% diff (EFLDA)** | 0.033 | -0.179 | 0.019 | 0.146 | 1.489 | -0.162 | -0.106 | 0.578 | -0.706 | 0.074 | -0.272 | 0.083 |
| | **% diff (TRA)** | 0.211 | 1.481 | 0.256 | 1.341 | 1.660 | 0.928 | 0.318 | 0.811 | -0.552 | 0.070 | 0.620 | 0.649 |
| **SVMSMOTE** | **EFLDA** | 0.990 | 0.961 | 0.983 | 0.960 | 0.942 | 0.981 | 0.994 | 0.931 | 0.991 | 0.973 | 0.980 | 0.971 |
| | **TRA** | 0.981 | 0.964 | 0.967 | 0.971 | 0.882 | 0.973 | 0.991 | 0.927 | 0.982 | 0.949 | 0.976 | 0.960 |
| | **% diff (EFLDA)** | 0.066 | -0.593 | 0.053 | -0.396 | 1.078 | 0.269 | 0.132 | -2.584 | -0.222 | -0.054 | 0.080 | -0.197 |
| | **% diff (TRA)** | -0.825 | -0.190 | -1.635 | 0.789 | -5.410 | -0.464 | -0.125 | -3.042 | -1.175 | -2.480 | -0.348 | -1.355 |
| **ROS** | **EFLDA** | 0.990 | 0.961 | 0.984 | 0.964 | 0.931 | 0.981 | 0.993 | 0.960 | 0.987 | 0.973 | 0.979 | 0.973 |
| | **TRA** | 0.989 | 0.984 | 0.982 | 0.983 | 0.937 | 0.984 | 0.995 | 0.958 | 0.987 | 0.968 | 0.985 | 0.977 |
| | **% diff (EFLDA)** | 0.106 | -0.500 | 0.120 | 0.003 | -0.095 | 0.306 | 0.106 | 0.411 | -0.683 | 0.013 | 0.004 | -0.019 |
| | **% diff (TRA)** | -0.030 | 1.864 | -0.139 | 1.961 | 0.509 | 0.653 | 0.289 | 0.202 | -0.675 | -0.567 | 0.588 | 0.423 |
| **RUS** | **EFLDA** | 0.989 | 0.965 | 0.982 | 0.960 | 0.935 | 0.977 | 0.991 | 0.960 | 0.986 | 0.972 | 0.982 | 0.973 |
| | **TRA** | 0.988 | 0.969 | 0.982 | 0.970 | 0.936 | 0.980 | 0.991 | 0.971 | 0.989 | 0.978 | 0.986 | 0.976 |
| | **% diff (EFLDA)** | -0.060 | -0.147 | -0.048 | -0.340 | 0.308 | -0.072 | -0.106 | 0.448 | -0.732 | -0.102 | 0.298 | -0.050 |
| | **% diff (TRA)** | -0.096 | 0.240 | -0.072 | 0.640 | 0.368 | 0.240 | -0.109 | 1.589 | -0.402 | 0.461 | 0.674 | 0.321 |
| **CC** | **EFLDA** | 0.990 | 0.966 | 0.982 | 0.960 | 0.933 | 0.976 | 0.992 | 0.955 | 0.984 | 0.974 | 0.978 | 0.972 |
| | **TRA** | 0.980 | 0.960 | 0.972 | 0.954 | 0.925 | 0.972 | 0.989 | 0.947 | 0.972 | 0.955 | 0.973 | 0.964 |
| | **% diff (EFLDA)** | 0.040 | 0.015 | -0.055 | -0.418 | 0.086 | -0.164 | -0.040 | -0.014 | -0.953 | 0.105 | -0.142 | -0.140 |
| | **% diff (TRA)** | -0.947 | -0.687 | -1.131 | -0.990 | -0.745 | -0.639 | -0.321 | -0.874 | -2.102 | -1.827 | -0.568 | -0.985 |

## Appendix Table 6 WBC Prognostics Results

| **WBC prognostics** | | | | | | | | | | | | | |
|---|---|---|---|---|---|---|---|---|---|---|---|---|---|
| | | **LR** | **KNN** | **SVM_L** | **SVM_R** | **DT** | **RF** | **NN** | **ADB** | **GB** | **NB** | **QDA** | **Avg** |
| **BEF** | **EFLDA** | 0.705 | 0.538 | 0.724 | 0.520 | 0.523 | 0.553 | 0.613 | 0.724 | 0.892 | 0.597 | 0.505 | 0.627 |
| | **TRA** | 0.705 | 0.538 | 0.724 | 0.520 | 0.523 | 0.553 | 0.613 | 0.724 | 0.892 | 0.597 | 0.505 | 0.627 |
| **ADASYN** | **EFLDA** | 0.710 | 0.536 | 0.667 | 0.535 | 0.473 | 0.661 | 0.633 | 0.695 | 0.898 | 0.614 | 0.492 | 0.629 |
| | **TRA** | 0.763 | 0.845 | 0.732 | 0.943 | 0.706 | 0.863 | 0.901 | 0.791 | 0.855 | 0.733 | 0.659 | 0.799 |
| | **% diff (EFLDA)** | 0.728 | -0.531 | -7.905 | 2.953 | -9.489 | 19.583 | 3.349 | -3.907 | 0.684 | 2.916 | -2.445 | 0.540 |
| | **% diff (TRA)** | 8.268 | 56.928 | 1.123 | 81.417 | 35.018 | 56.161 | 47.072 | 9.288 | -4.183 | 22.845 | 30.622 | 31.324 |
| **SMOTE** | **EFLDA** | 0.712 | 0.528 | 0.674 | 0.529 | 0.577 | 0.592 | 0.670 | 0.671 | 0.910 | 0.625 | 0.517 | 0.637 |
| | **TRA** | 0.764 | 0.825 | 0.728 | 0.942 | 0.605 | 0.873 | 0.901 | 0.733 | 0.907 | 0.733 | 0.665 | 0.789 |
| | **% diff (EFLDA)** | 0.982 | -1.902 | -6.912 | 1.818 | 10.315 | 7.082 | 9.404 | -7.271 | 2.021 | 4.718 | 2.460 | 2.065 |
| | **% diff (TRA)** | 8.349 | 53.163 | 0.551 | 81.187 | 15.728 | 57.957 | 47.061 | 1.324 | 1.616 | 22.751 | 31.774 | 29.224 |
| **SVMSMOTE** | **EFLDA** | 0.726 | 0.529 | 0.701 | 0.548 | 0.503 | 0.613 | 0.649 | 0.707 | 0.898 | 0.616 | 0.499 | 0.635 |
| | **TRA** | 0.770 | 0.762 | 0.724 | 0.883 | 0.633 | 0.832 | 0.859 | 0.752 | 0.895 | 0.726 | 0.645 | 0.771 |
| | **% diff (EFLDA)** | 2.894 | -1.743 | -3.206 | 5.410 | -3.756 | 10.864 | 5.951 | -2.243 | 0.644 | 3.190 | -1.151 | 1.532 |
| | **% diff (TRA)** | 9.241 | 41.506 | -0.085 | 69.844 | 21.198 | 50.671 | 40.223 | 3.957 | 0.280 | 21.556 | 27.728 | 26.011 |
| **ROS** | **EFLDA** | 0.717 | 0.543 | 0.676 | 0.526 | 0.530 | 0.572 | 0.622 | 0.667 | 0.894 | 0.599 | 0.484 | 0.621 |
| | **TRA** | 0.776 | 0.766 | 0.754 | 0.983 | 0.624 | 0.917 | 0.874 | 0.723 | 0.887 | 0.694 | 0.634 | 0.785 |
| | **% diff (EFLDA)** | 1.728 | 0.863 | -6.684 | 1.268 | 1.334 | 3.532 | 1.593 | -7.779 | 0.203 | 0.315 | -4.049 | -0.698 |
| | **% diff (TRA)** | 10.034 | 42.185 | 4.096 | 89.083 | 19.473 | 66.045 | 42.654 | -0.123 | -0.549 | 16.205 | 25.532 | 28.603 |
| **RUS** | **EFLDA** | 0.736 | 0.558 | 0.683 | 0.397 | 0.614 | 0.654 | 0.727 | 0.699 | 0.873 | 0.628 | 0.633 | 0.655 |
| | **TRA** | 0.837 | 0.669 | 0.789 | 0.358 | 0.624 | 0.668 | 0.700 | 0.796 | 0.913 | 0.714 | 0.546 | 0.692 |
| | **% diff (EFLDA)** | 4.325 | 3.660 | -5.612 | -23.683 | 17.421 | 18.342 | 18.632 | -3.411 | -2.144 | 5.218 | 25.350 | 5.282 |
| | **% diff (TRA)** | 18.675 | 24.149 | 8.942 | -31.105 | 19.339 | 20.928 | 14.202 | 10.019 | 2.376 | 19.543 | 8.103 | 10.470 |
| **CC** | **EFLDA** | 0.712 | 0.528 | 0.704 | 0.492 | 0.560 | 0.556 | 0.626 | 0.601 | 0.845 | 0.671 | 0.496 | 0.617 |
| | **TRA** | 0.765 | 0.713 | 0.764 | 0.354 | 0.623 | 0.760 | 0.719 | 0.675 | 0.831 | 0.676 | 0.613 | 0.681 |
| | **% diff (EFLDA)** | 1.014 | -2.025 | -2.847 | -5.297 | 7.121 | 0.687 | 2.152 | -16.886 | -5.251 | 12.373 | -1.713 | -0.970 |
| | **% diff (TRA)** | 8.485 | 32.449 | 5.532 | -31.984 | 19.221 | 37.530 | 17.387 | -6.682 | -6.868 | 13.297 | 21.387 | 9.978 |

## Appendix Table 7 SEER results

| **SEER** | | | | | | | | | | | | | |
|---|---|---|---|---|---|---|---|---|---|---|---|---|---|
| | | **LR** | **KNN** | **SVM_L** | **SVM_R** | **DT** | **RF** | **NN** | **ADB** | **GB** | **NB** | **QDA** | **Avg** |
| **BEF** | **EFLDA** | 0.966 | 0.745 | 0.975 | 0.607 | 0.742 | 0.648 | 0.902 | 0.637 | 0.908 | 0.846 | 0.853 | 0.803 |
| | **TRA** | 0.966 | 0.745 | 0.975 | 0.607 | 0.742 | 0.648 | 0.902 | 0.637 | 0.908 | 0.846 | 0.853 | 0.803 |
| **ADASYN** | **EFLDA** | 0.911 | 0.705 | 0.912 | 0.590 | 0.703 | 0.635 | 0.766 | 0.618 | 0.894 | 0.813 | 0.805 | 0.759 |
| | **TRA** | 0.879 | 0.808 | 0.878 | 0.780 | 0.715 | 0.635 | 0.765 | 0.518 | 0.842 | 0.786 | 0.809 | 0.765 |
| | **% diff (EFLDA)** | -5.694 | -5.262 | -6.479 | -2.738 | -5.307 | -1.938 | -15.094 | -2.911 | -1.630 | -3.877 | -5.614 | -5.140 |
| | **% diff (TRA)** | -9.043 | 8.576 | -9.918 | 28.582 | -3.630 | -1.949 | -15.129 | -18.710 | -7.311 | -7.048 | -5.096 | -3.698 |
| **SMOTE** | **EFLDA** | 0.939 | 0.722 | 0.941 | 0.619 | 0.750 | 0.626 | 0.830 | 0.638 | 0.921 | 0.832 | 0.828 | 0.786 |
| | **TRA** | 0.873 | 0.900 | 0.872 | 0.903 | 0.733 | 0.788 | 0.871 | 0.705 | 0.853 | 0.814 | 0.832 | 0.831 |
| | **% diff (EFLDA)** | -2.755 | -3.094 | -3.482 | 1.959 | 1.055 | -3.377 | -8.022 | 0.222 | 1.434 | -1.657 | -2.924 | -1.876 |
| | **% diff (TRA)** | -9.677 | 20.842 | -10.530 | 48.779 | -1.255 | 21.617 | -3.411 | 10.749 | -6.046 | -3.739 | -2.397 | 5.903 |
| **SVMSMOTE** | **EFLDA** | 0.965 | 0.714 | 0.967 | 0.604 | 0.676 | 0.687 | 0.774 | 0.657 | 0.913 | 0.863 | 0.869 | 0.790 |
| | **TRA** | 0.871 | 0.904 | 0.870 | 0.913 | 0.732 | 0.837 | 0.882 | 0.757 | 0.854 | 0.822 | 0.839 | 0.844 |
| | **% diff (EFLDA)** | -0.080 | -4.084 | -0.785 | -0.493 | -8.906 | 6.046 | -14.155 | 3.088 | 0.514 | 1.980 | 1.843 | -1.366 |
| | **% diff (TRA)** | -9.865 | 21.357 | -10.790 | 50.450 | -1.378 | 29.200 | -2.232 | 18.817 | -5.973 | -2.861 | -1.624 | 7.736 |
| **ROS** | **EFLDA** | 0.965 | 0.705 | 0.966 | 0.602 | 0.756 | 0.643 | 0.838 | 0.645 | 0.923 | 0.845 | 0.853 | 0.795 |
| | **TRA** | 0.855 | 0.906 | 0.855 | 0.991 | 0.691 | 0.716 | 0.860 | 0.688 | 0.865 | 0.784 | 0.794 | 0.819 |
| | **% diff (EFLDA)** | -0.086 | -5.323 | -0.902 | -0.773 | 1.933 | -0.759 | -7.086 | 1.290 | 1.642 | -0.141 | 0.005 | -0.927 |
| | **% diff (TRA)** | -11.473 | 21.687 | -12.281 | 63.337 | -6.864 | 10.533 | -4.671 | 8.084 | -4.757 | -7.280 | -6.892 | 4.493 |
| **RUS** | **EFLDA** | 0.968 | 0.784 | 0.968 | 0.608 | 0.755 | 0.703 | 0.865 | 0.631 | 0.920 | 0.847 | 0.854 | 0.809 |
| | **TRA** | 0.842 | 0.775 | 0.847 | 0.715 | 0.729 | 0.762 | 0.791 | 0.665 | 0.868 | 0.781 | 0.785 | 0.778 |
| | **% diff (EFLDA)** | 0.162 | 5.349 | -0.695 | 0.160 | 1.821 | 8.627 | -4.086 | -0.871 | 1.303 | 0.171 | 0.085 | 1.093 |
| | **% diff (TRA)** | -12.828 | 4.057 | -13.114 | 17.767 | -1.749 | 17.678 | -12.249 | 4.377 | -4.450 | -7.621 | -7.942 | -1.461 |
| **CC** | **EFLDA** | 0.832 | 0.634 | 0.852 | 0.487 | 0.624 | 0.542 | 0.755 | 0.471 | 0.916 | 0.767 | 0.730 | 0.692 |
| | **TRA** | 0.855 | 0.871 | 0.853 | 0.788 | 0.733 | 0.818 | 0.869 | 0.674 | 0.830 | 0.816 | 0.855 | 0.815 |
| | **% diff (EFLDA)** | -13.854 | -14.783 | -12.548 | -19.767 | -15.844 | -16.341 | -16.266 | -26.055 | 0.878 | -9.353 | -14.422 | -14.396 |
| | **% diff (TRA)** | -11.525 | 16.994 | -12.469 | 29.951 | -1.204 | 26.276 | -3.680 | 5.895 | -8.662 | -3.564 | 0.243 | 3.478 |

## Appendix Table 8 SegmentPCP_4 Results

| **SegmentPCP4** | | | | | | | | | | | | | |
|---|---|---|---|---|---|---|---|---|---|---|---|---|---|
| | | **LR** | **KNN** | **SVM_L** | **SVM_R** | **DT** | **RF** | **NN** | **ADB** | **GB** | **NB** | **QDA** | **Avg** |
| **BEF** | **EFLDA** | 0.722 | 0.668 | 0.579 | 0.512 | 0.618 | 0.664 | 0.663 | 0.655 | 0.860 | 0.661 | 0.500 | 0.646 |
| | **TRA** | 0.722 | 0.668 | 0.579 | 0.512 | 0.618 | 0.664 | 0.663 | 0.655 | 0.860 | 0.661 | 0.500 | 0.646 |
| **ADASYN** | **EFLDA** | 0.717 | 0.689 | 0.714 | 0.515 | 0.633 | 0.721 | 0.639 | 0.671 | 0.824 | 0.655 | 0.500 | 0.662 |
| | **TRA** | 0.734 | 0.940 | 0.733 | 0.855 | 0.701 | 0.848 | 0.912 | 0.743 | 0.773 | 0.660 | 0.500 | 0.764 |
| | **% diff (EFLDA)** | -0.657 | 3.057 | 23.284 | 0.560 | 2.477 | 8.668 | -3.745 | 2.431 | -4.175 | -0.926 | 0.023 | 2.818 |
| | **% diff (TRA)** | 1.709 | 40.622 | 26.614 | 67.001 | 13.407 | 27.751 | 37.455 | 13.478 | -10.152 | -0.137 | 0.023 | 19.797 |
| **SMOTE** | **EFLDA** | 0.718 | 0.689 | 0.716 | 0.515 | 0.620 | 0.710 | 0.622 | 0.681 | 0.822 | 0.655 | 0.500 | 0.659 |
| | **TRA** | 0.753 | 0.963 | 0.753 | 0.949 | 0.719 | 0.886 | 0.940 | 0.750 | 0.784 | 0.669 | 0.500 | 0.788 |
| | **% diff (EFLDA)** | -0.484 | 3.075 | 23.643 | 0.560 | 0.249 | 6.945 | -6.264 | 4.066 | -4.469 | -0.964 | 0.023 | 2.398 |
| | **% diff (TRA)** | 4.360 | 44.085 | 29.992 | 85.199 | 16.347 | 33.465 | 41.674 | 14.609 | -8.829 | 1.177 | 0.023 | 23.827 |
| **SVMSMOTE** | **EFLDA** | 0.714 | 0.693 | 0.713 | 0.521 | 0.636 | 0.704 | 0.662 | 0.674 | 0.792 | 0.658 | 0.500 | 0.661 |
| | **TRA** | 0.813 | 0.967 | 0.808 | 0.924 | 0.770 | 0.903 | 0.928 | 0.762 | 0.828 | 0.733 | 0.500 | 0.812 |
| | **% diff (EFLDA)** | -1.103 | 3.677 | 23.200 | 1.756 | 2.898 | 6.046 | -0.157 | 2.870 | -7.973 | -0.521 | 0.023 | 2.793 |
| | **% diff (TRA)** | 12.632 | 44.743 | 39.556 | 80.340 | 24.591 | 36.031 | 39.952 | 16.384 | -3.701 | 10.903 | 0.023 | 27.405 |
| **ROS** | **EFLDA** | 0.718 | 0.666 | 0.718 | 0.511 | 0.620 | 0.704 | 0.666 | 0.691 | 0.809 | 0.661 | 0.500 | 0.660 |
| | **TRA** | 0.739 | 0.981 | 0.738 | 1.000 | 0.703 | 0.918 | 0.926 | 0.719 | 0.801 | 0.663 | 0.500 | 0.790 |
| | **% diff (EFLDA)** | -0.527 | -0.390 | 23.986 | -0.180 | 0.335 | 6.113 | 0.372 | 5.474 | -5.972 | 0.027 | 0.023 | 2.660 |
| | **% diff (TRA)** | 2.457 | 46.794 | 27.513 | 95.254 | 13.802 | 38.353 | 39.587 | 9.827 | -6.836 | 0.293 | 0.023 | 24.279 |
| **RUS** | **EFLDA** | 0.705 | 0.662 | 0.709 | 0.496 | 0.629 | 0.677 | 0.676 | 0.662 | 0.812 | 0.657 | 0.510 | 0.654 |
| | **TRA** | 0.713 | 0.661 | 0.709 | 0.496 | 0.617 | 0.639 | 0.644 | 0.684 | 0.818 | 0.670 | 0.502 | 0.650 |
| | **% diff (EFLDA)** | -2.308 | -0.899 | 22.492 | -3.094 | 1.752 | 2.061 | 1.920 | 1.177 | -5.594 | -0.664 | 1.982 | 1.711 |
| | **% diff (TRA)** | -1.142 | -1.070 | 22.403 | -3.117 | -0.149 | -3.642 | -2.909 | 4.406 | -4.888 | 1.355 | 0.495 | 1.068 |
| **CC** | **EFLDA** | 0.696 | 0.668 | 0.695 | 0.496 | 0.540 | 0.606 | 0.603 | 0.576 | 0.795 | 0.639 | 0.497 | 0.619 |
| | **TRA** | 0.746 | 0.835 | 0.747 | 0.496 | 0.729 | 0.904 | 0.879 | 0.840 | 0.834 | 0.794 | 0.515 | 0.756 |
| | **% diff (EFLDA)** | -3.556 | 0.009 | 20.056 | -3.117 | -12.589 | -8.753 | -9.155 | -12.105 | -7.633 | -3.334 | -0.625 | -3.709 |
| | **% diff (TRA)** | 3.332 | 25.004 | 29.049 | -3.117 | 17.908 | 36.244 | 32.578 | 28.249 | -3.090 | 20.113 | 3.035 | 17.210 |

## Appendix Table 9 SegmentPCP_5 Results

| **SegmentPCP5** | | | | | | | | | | | | | |
|---|---|---|---|---|---|---|---|---|---|---|---|---|---|
| | | **LR** | **KNN** | **SVM_L** | **SVM_R** | **DT** | **RF** | **NN** | **ADB** | **GB** | **NB** | **QDA** | **Avg** |
| **BEF** | **EFLDA** | 0.724 | 0.604 | 0.537 | 0.512 | 0.680 | 0.650 | 0.646 | 0.661 | 0.876 | 0.681 | 0.500 | 0.643 |
| | **TRA** | 0.724 | 0.604 | 0.537 | 0.512 | 0.680 | 0.650 | 0.646 | 0.661 | 0.876 | 0.681 | 0.500 | 0.643 |
| **ADASYN** | **EFLDA** | 0.704 | 0.638 | 0.712 | 0.488 | 0.662 | 0.695 | 0.668 | 0.674 | 0.842 | 0.677 | 0.500 | 0.660 |
| | **TRA** | 0.729 | 0.923 | 0.735 | 0.870 | 0.749 | 0.869 | 0.955 | 0.740 | 0.789 | 0.720 | 0.500 | 0.780 |
| | **% diff (EFLDA)** | -2.648 | 5.677 | 32.630 | -4.649 | -2.548 | 6.968 | 3.338 | 1.923 | -3.840 | -0.493 | 0.000 | 3.305 |
| | **% diff (TRA)** | 0.809 | 52.952 | 36.871 | 70.004 | 10.212 | 33.686 | 47.725 | 11.903 | -9.861 | 5.729 | 0.000 | 23.639 |
| **SMOTE** | **EFLDA** | 0.708 | 0.629 | 0.713 | 0.489 | 0.668 | 0.691 | 0.658 | 0.681 | 0.834 | 0.681 | 0.500 | 0.659 |
| | **TRA** | 0.767 | 0.942 | 0.768 | 0.977 | 0.780 | 0.924 | 0.969 | 0.797 | 0.798 | 0.733 | 0.500 | 0.814 |
| | **% diff (EFLDA)** | -2.116 | 4.157 | 32.699 | -4.504 | -1.760 | 6.346 | 1.757 | 3.039 | -4.802 | 0.089 | 0.000 | 3.173 |
| | **% diff (TRA)** | 6.000 | 56.128 | 43.079 | 90.939 | 14.836 | 42.240 | 49.894 | 20.502 | -8.891 | 7.629 | 0.000 | 29.305 |
| **SVMSMOTE** | **EFLDA** | 0.718 | 0.630 | 0.714 | 0.489 | 0.673 | 0.686 | 0.652 | 0.685 | 0.845 | 0.684 | 0.500 | 0.662 |
| | **TRA** | 0.836 | 0.961 | 0.832 | 0.959 | 0.808 | 0.941 | 0.972 | 0.823 | 0.844 | 0.777 | 0.500 | 0.841 |
| | **% diff (EFLDA)** | -0.746 | 4.364 | 33.037 | -4.447 | -0.916 | 5.513 | 0.922 | 3.574 | -3.456 | 0.534 | 0.000 | 3.489 |
| | **% diff (TRA)** | 15.490 | 59.277 | 54.936 | 87.350 | 18.890 | 44.722 | 50.461 | 24.423 | -3.668 | 14.128 | 0.000 | 33.274 |
| **ROS** | **EFLDA** | 0.713 | 0.601 | 0.721 | 0.486 | 0.676 | 0.707 | 0.656 | 0.686 | 0.837 | 0.677 | 0.500 | 0.660 |
| | **TRA** | 0.757 | 0.971 | 0.757 | 0.999 | 0.783 | 0.958 | 0.966 | 0.771 | 0.833 | 0.692 | 0.500 | 0.817 |
| | **% diff (EFLDA)** | -1.498 | -0.451 | 34.327 | -5.077 | -0.531 | 8.762 | 1.486 | 3.781 | -4.407 | -0.481 | 0.000 | 3.265 |
| | **% diff (TRA)** | 4.561 | 60.907 | 40.875 | 95.079 | 15.235 | 47.374 | 49.461 | 16.654 | -4.900 | 1.731 | 0.000 | 29.725 |
| **RUS** | **EFLDA** | 0.676 | 0.645 | 0.672 | 0.500 | 0.630 | 0.676 | 0.668 | 0.667 | 0.821 | 0.679 | 0.513 | 0.650 |
| | **TRA** | 0.714 | 0.677 | 0.712 | 0.492 | 0.642 | 0.724 | 0.704 | 0.658 | 0.822 | 0.708 | 0.504 | 0.669 |
| | **% diff (EFLDA)** | -6.548 | 6.809 | 25.219 | -2.293 | -7.318 | 3.933 | 3.379 | 0.853 | -6.256 | -0.232 | 2.660 | 1.837 |
| | **% diff (TRA)** | -1.362 | 12.115 | 32.669 | -3.823 | -5.541 | 11.350 | 8.910 | -0.411 | -6.112 | 3.967 | 0.769 | 4.776 |
| **CC** | **EFLDA** | 0.673 | 0.669 | 0.696 | 0.501 | 0.618 | 0.620 | 0.637 | 0.586 | 0.822 | 0.685 | 0.508 | 0.638 |
| | **TRA** | 0.762 | 0.788 | 0.767 | 0.500 | 0.705 | 0.908 | 0.842 | 0.828 | 0.859 | 0.736 | 0.507 | 0.746 |
| | **% diff (EFLDA)** | -6.952 | 10.785 | 29.531 | -2.071 | -9.018 | -4.645 | -1.451 | -11.293 | -6.124 | 0.639 | 1.679 | 0.098 |
| | **% diff (TRA)** | 5.263 | 30.632 | 42.877 | -2.321 | 3.700 | 39.637 | 30.317 | 25.285 | -1.935 | 8.093 | 1.396 | 16.631 |

## Appendix Table 10 SegmentPCP_6 Results

| **SegmentPCP6** | | | | | | | | | | | | | |
|---|---|---|---|---|---|---|---|---|---|---|---|---|---|
| | | **LR** | **KNN** | **SVM_L** | **SVM_R** | **DT** | **RF** | **NN** | **ADB** | **GB** | **NB** | **QDA** | **Avg** |
| **BEF** | **EFLDA** | 0.526 | 0.508 | 0.514 | 0.524 | 0.462 | 0.611 | 0.442 | 0.636 | 0.867 | 0.494 | 0.477 | 0.551 |
| | **TRA** | 0.526 | 0.508 | 0.514 | 0.524 | 0.462 | 0.611 | 0.442 | 0.636 | 0.867 | 0.494 | 0.477 | 0.551 |
| **ADASYN** | **EFLDA** | 0.480 | 0.516 | 0.487 | 0.474 | 0.481 | 0.539 | 0.442 | 0.524 | 0.877 | 0.498 | 0.498 | 0.529 |
| | **TRA** | 0.660 | 0.888 | 0.660 | 0.855 | 0.729 | 0.948 | 0.950 | 0.765 | 0.787 | 0.689 | 0.998 | 0.812 |
| | **% diff (EFLDA)** | -8.785 | 1.711 | -5.182 | -9.481 | 4.216 | -11.746 | 0.073 | -17.588 | 1.189 | 0.878 | 4.346 | -3.670 |
| | **% diff (TRA)** | 25.484 | 74.931 | 28.410 | 63.219 | 57.716 | 55.160 | 115.041 | 20.237 | -9.217 | 39.430 | 109.185 | 52.691 |
| **SMOTE** | **EFLDA** | 0.496 | 0.510 | 0.515 | 0.473 | 0.506 | 0.554 | 0.462 | 0.526 | 0.860 | 0.499 | 0.498 | 0.536 |
| | **TRA** | 0.724 | 0.909 | 0.729 | 0.944 | 0.783 | 0.970 | 0.959 | 0.805 | 0.773 | 0.742 | 0.998 | 0.849 |
| | **% diff (EFLDA)** | -5.665 | 0.442 | 0.320 | -9.615 | 9.555 | -9.364 | 4.563 | -17.246 | -0.859 | 1.076 | 4.346 | -2.041 |
| | **% diff (TRA)** | 37.717 | 79.027 | 41.895 | 80.193 | 69.593 | 58.889 | 117.094 | 26.541 | -10.816 | 50.286 | 109.185 | 59.964 |
| **SVMSMOTE** | **EFLDA** | 0.527 | 0.488 | 0.577 | 0.472 | 0.474 | 0.522 | 0.447 | 0.553 | 0.858 | 0.493 | 0.498 | 0.537 |
| | **TRA** | 0.757 | 0.925 | 0.757 | 0.921 | 0.830 | 0.958 | 0.958 | 0.794 | 0.824 | 0.750 | 0.988 | 0.860 |
| | **% diff (EFLDA)** | 0.295 | -3.915 | 12.298 | -9.949 | 2.574 | -14.561 | 1.209 | -13.072 | -1.107 | -0.291 | 4.346 | -2.016 |
| | **% diff (TRA)** | 44.031 | 82.258 | 47.311 | 75.806 | 79.743 | 56.797 | 116.785 | 24.838 | -4.992 | 51.793 | 107.219 | 61.963 |
| **ROS** | **EFLDA** | 0.496 | 0.492 | 0.489 | 0.474 | 0.534 | 0.541 | 0.447 | 0.608 | 0.850 | 0.510 | 0.498 | 0.540 |
| | **TRA** | 0.700 | 0.939 | 0.692 | 0.998 | 0.847 | 0.993 | 0.965 | 0.794 | 0.858 | 0.645 | 0.998 | 0.857 |
| | **% diff (EFLDA)** | -5.582 | -3.079 | -4.732 | -9.451 | 15.543 | -11.358 | 1.142 | -4.415 | -2.011 | 3.260 | 4.346 | -1.485 |
| | **% diff (TRA)** | 33.222 | 84.975 | 34.657 | 90.542 | 83.418 | 62.554 | 118.415 | 24.885 | -1.104 | 30.515 | 109.185 | 61.024 |
| **RUS** | **EFLDA** | 0.467 | 0.509 | 0.487 | 0.500 | 0.481 | 0.496 | 0.425 | 0.476 | 0.836 | 0.439 | 0.497 | 0.510 |
| | **TRA** | 0.503 | 0.619 | 0.475 | 0.500 | 0.529 | 0.612 | 0.516 | 0.526 | 0.895 | 0.607 | 0.448 | 0.566 |
| | **% diff (EFLDA)** | -11.210 | 0.329 | -5.268 | -4.549 | 4.035 | -18.783 | -3.895 | -25.110 | -3.617 | -11.184 | 4.144 | -6.828 |
| | **% diff (TRA)** | -4.273 | 21.935 | -7.575 | -4.550 | 14.524 | 0.157 | 16.678 | -17.391 | 3.170 | 22.896 | -6.019 | 3.595 |
| **CC** | **EFLDA** | 0.443 | 0.515 | 0.471 | 0.501 | 0.536 | 0.582 | 0.529 | 0.575 | 0.847 | 0.411 | 0.501 | 0.537 |
| | **TRA** | 0.400 | 0.552 | 0.423 | 0.500 | 0.572 | 0.541 | 0.592 | 0.535 | 0.908 | 0.508 | 0.563 | 0.554 |
| | **% diff (EFLDA)** | -15.776 | 1.447 | -8.396 | -4.399 | 16.056 | -4.775 | 19.651 | -9.645 | -2.309 | -16.874 | 5.037 | -1.817 |
| | **% diff (TRA)** | -23.923 | 8.766 | -17.707 | -4.550 | 23.801 | -11.371 | 33.940 | -15.827 | 4.670 | 2.861 | 17.944 | 1.691 |

## Appendix Table 11 Pima-Diabetes Results

| **Pima-Diabetes** | | | | | | | | | | | | | |
|---|---|---|---|---|---|---|---|---|---|---|---|---|---|
| | | **LR** | **KNN** | **SVM_L** | **SVM_R** | **DT** | **RF** | **NN** | **ADB** | **GB** | **NB** | **QDA** | **Avg** |
| **BEF** | **EFLDA** | 0.833 | 0.782 | 0.834 | 0.744 | 0.696 | 0.826 | 0.834 | 0.769 | 0.920 | 0.817 | 0.812 | 0.806 |
| | **TRA** | 0.833 | 0.782 | 0.834 | 0.744 | 0.696 | 0.826 | 0.834 | 0.769 | 0.920 | 0.817 | 0.812 | 0.806 |
| **ADASYN** | **EFLDA** | 0.828 | 0.769 | 0.830 | 0.742 | 0.693 | 0.810 | 0.794 | 0.783 | 0.913 | 0.817 | 0.812 | 0.799 |
| | **TRA** | 0.807 | 0.818 | 0.804 | 0.879 | 0.676 | 0.812 | 0.814 | 0.757 | 0.880 | 0.775 | 0.785 | 0.801 |
| | **% diff (EFLDA)** | -0.554 | -1.699 | -0.374 | -0.324 | -0.408 | -1.907 | -4.810 | 1.840 | -0.754 | -0.042 | -0.096 | -0.830 |
| | **% diff (TRA)** | -3.165 | 4.579 | -3.500 | 18.096 | -2.917 | -1.699 | -2.448 | -1.605 | -4.329 | -5.166 | -3.381 | -0.503 |
| **SMOTE** | **EFLDA** | 0.834 | 0.791 | 0.833 | 0.757 | 0.636 | 0.829 | 0.805 | 0.776 | 0.909 | 0.821 | 0.815 | 0.800 |
| | **TRA** | 0.842 | 0.846 | 0.841 | 0.890 | 0.703 | 0.852 | 0.852 | 0.800 | 0.907 | 0.820 | 0.824 | 0.834 |
| | **% diff (EFLDA)** | 0.101 | 1.140 | -0.071 | 1.754 | -8.617 | 0.314 | -3.422 | 0.834 | -1.199 | 0.435 | 0.313 | -0.765 |
| | **% diff (TRA)** | 1.037 | 8.161 | 0.940 | 19.619 | 1.057 | 3.202 | 2.204 | 3.989 | -1.421 | 0.350 | 1.431 | 3.688 |
| **SVMSMOTE** | **EFLDA** | 0.833 | 0.786 | 0.832 | 0.755 | 0.657 | 0.816 | 0.814 | 0.778 | 0.906 | 0.815 | 0.816 | 0.801 |
| | **TRA** | 0.845 | 0.842 | 0.846 | 0.888 | 0.695 | 0.857 | 0.851 | 0.800 | 0.911 | 0.831 | 0.826 | 0.835 |
| | **% diff (EFLDA)** | 0.019 | 0.504 | -0.131 | 1.437 | -5.628 | -1.203 | -2.425 | 1.151 | -1.490 | -0.281 | 0.452 | -0.690 |
| | **% diff (TRA)** | 1.438 | 7.633 | 1.441 | 19.269 | -0.169 | 3.727 | 1.991 | 4.080 | -0.976 | 1.650 | 1.677 | 3.796 |
| **ROS** | **EFLDA** | 0.830 | 0.790 | 0.830 | 0.748 | 0.672 | 0.821 | 0.818 | 0.779 | 0.901 | 0.815 | 0.809 | 0.801 |
| | **TRA** | 0.842 | 0.847 | 0.842 | 0.935 | 0.706 | 0.874 | 0.863 | 0.804 | 0.903 | 0.829 | 0.832 | 0.843 |
| | **% diff (EFLDA)** | -0.313 | 0.920 | -0.414 | 0.489 | -3.464 | -0.571 | -1.886 | 1.226 | -2.081 | -0.245 | -0.370 | -0.610 |
| | **% diff (TRA)** | 1.057 | 8.267 | 0.970 | 25.559 | 1.387 | 5.772 | 3.424 | 4.520 | -1.770 | 1.444 | 2.362 | 4.818 |
| **RUS** | **EFLDA** | 0.829 | 0.785 | 0.830 | 0.748 | 0.677 | 0.817 | 0.823 | 0.793 | 0.906 | 0.808 | 0.806 | 0.802 |
| | **TRA** | 0.834 | 0.784 | 0.834 | 0.742 | 0.650 | 0.822 | 0.829 | 0.788 | 0.914 | 0.812 | 0.807 | 0.801 |
| | **% diff (EFLDA)** | -0.526 | 0.343 | -0.411 | 0.493 | -2.704 | -1.091 | -1.331 | 3.057 | -1.468 | -1.131 | -0.793 | -0.506 |
| | **% diff (TRA)** | 0.190 | 0.205 | 0.051 | -0.328 | -6.554 | -0.517 | -0.565 | 2.413 | -0.667 | -0.685 | -0.684 | -0.649 |
| **CC** | **EFLDA** | 0.825 | 0.772 | 0.829 | 0.737 | 0.668 | 0.810 | 0.817 | 0.806 | 0.896 | 0.817 | 0.800 | 0.798 |
| | **TRA** | 0.794 | 0.751 | 0.799 | 0.705 | 0.662 | 0.772 | 0.783 | 0.772 | 0.867 | 0.773 | 0.766 | 0.768 |
| | **% diff (EFLDA)** | -0.920 | -1.312 | -0.567 | -0.975 | -3.950 | -1.986 | -2.064 | 4.859 | -2.629 | 0.028 | -1.560 | -1.007 |
| | **% diff (TRA)** | -4.680 | -3.965 | -4.172 | -5.283 | -4.939 | -6.534 | -6.151 | 0.328 | -5.776 | -5.366 | -5.660 | -4.745 |

# Result Figures

## Appendix Figure  1 LSM_5Year - TRA

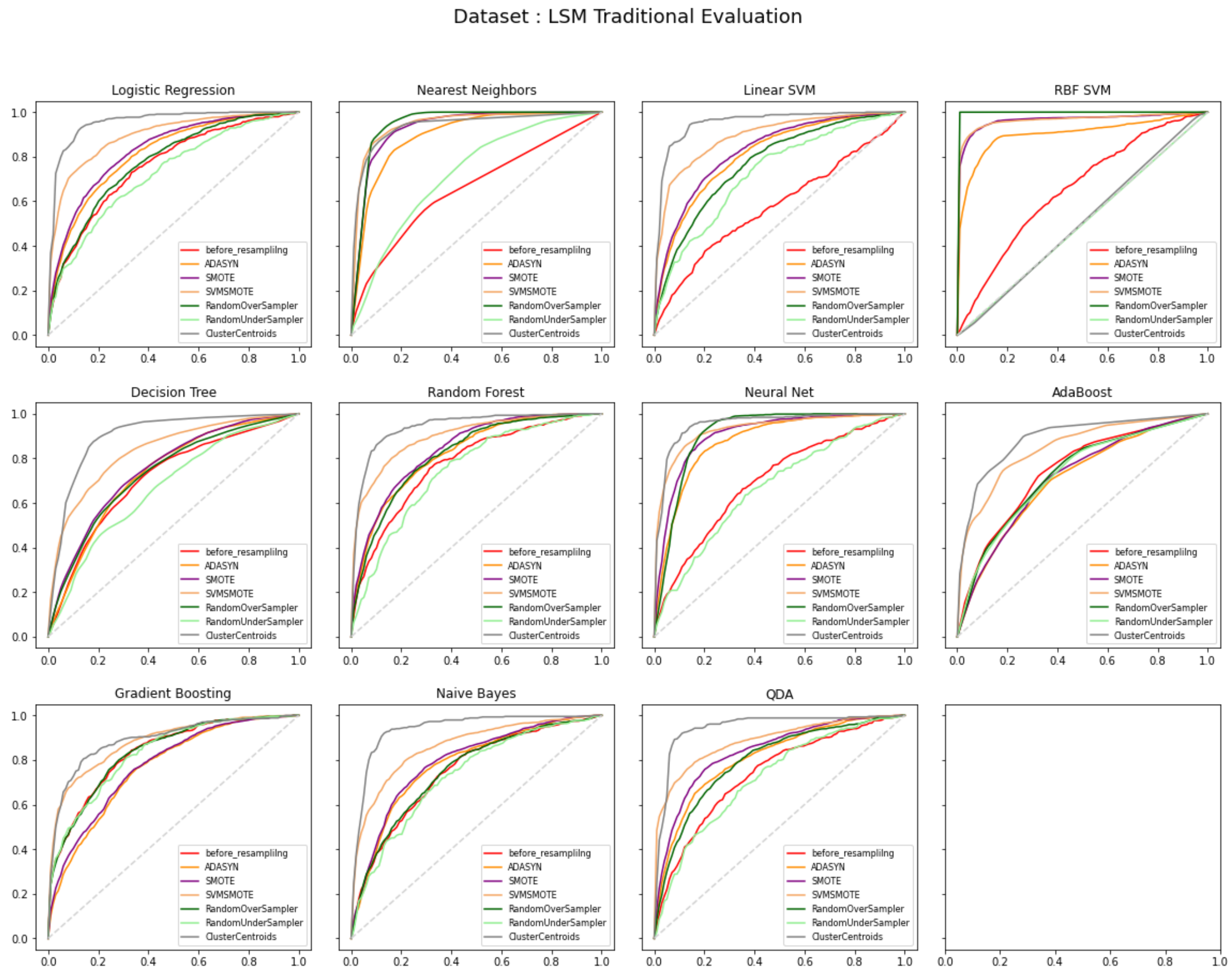

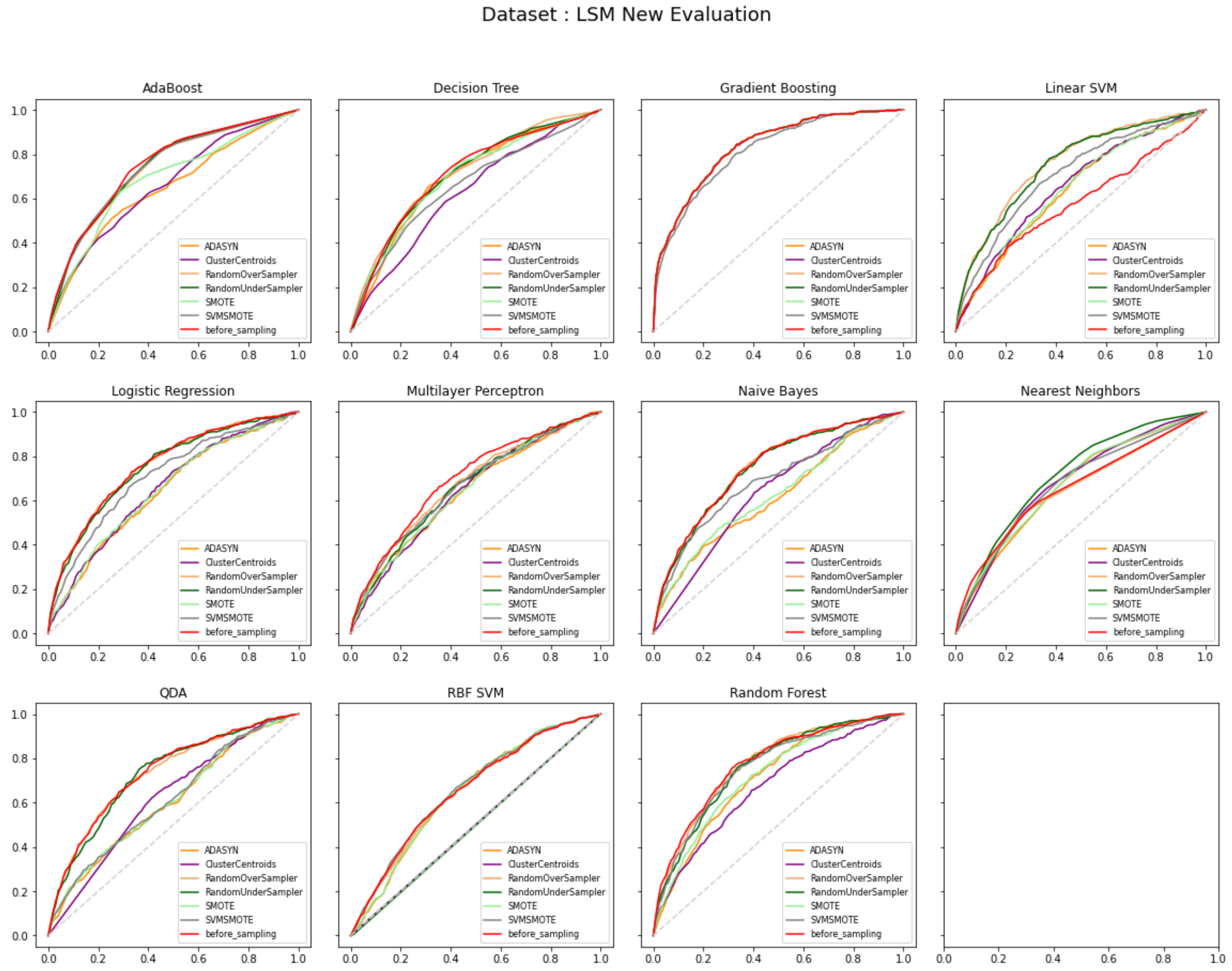

**Appendix Figure 2 LSM-5 Year EFLDA**

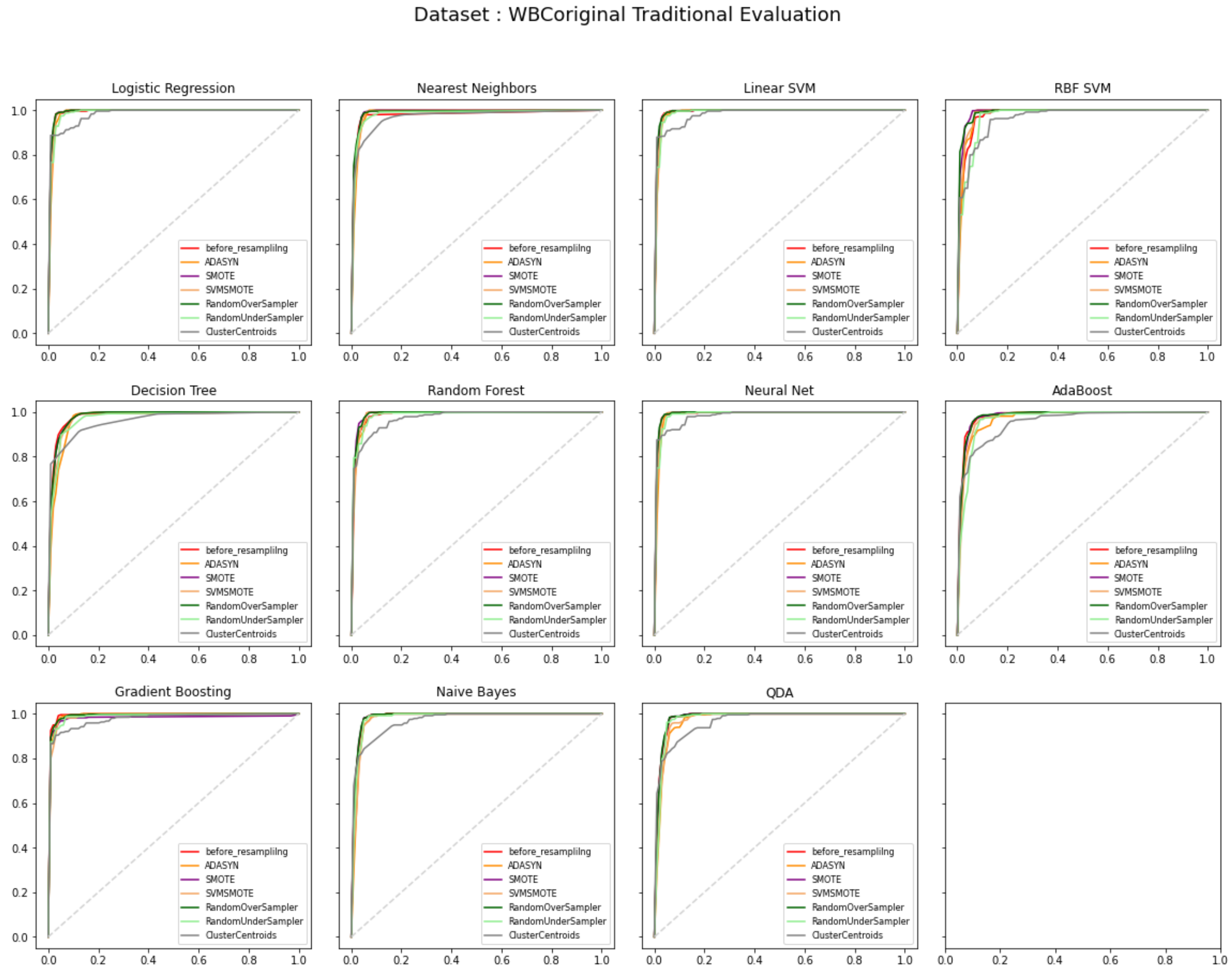

**Appendix Figure 3 WBC Original -TRA**

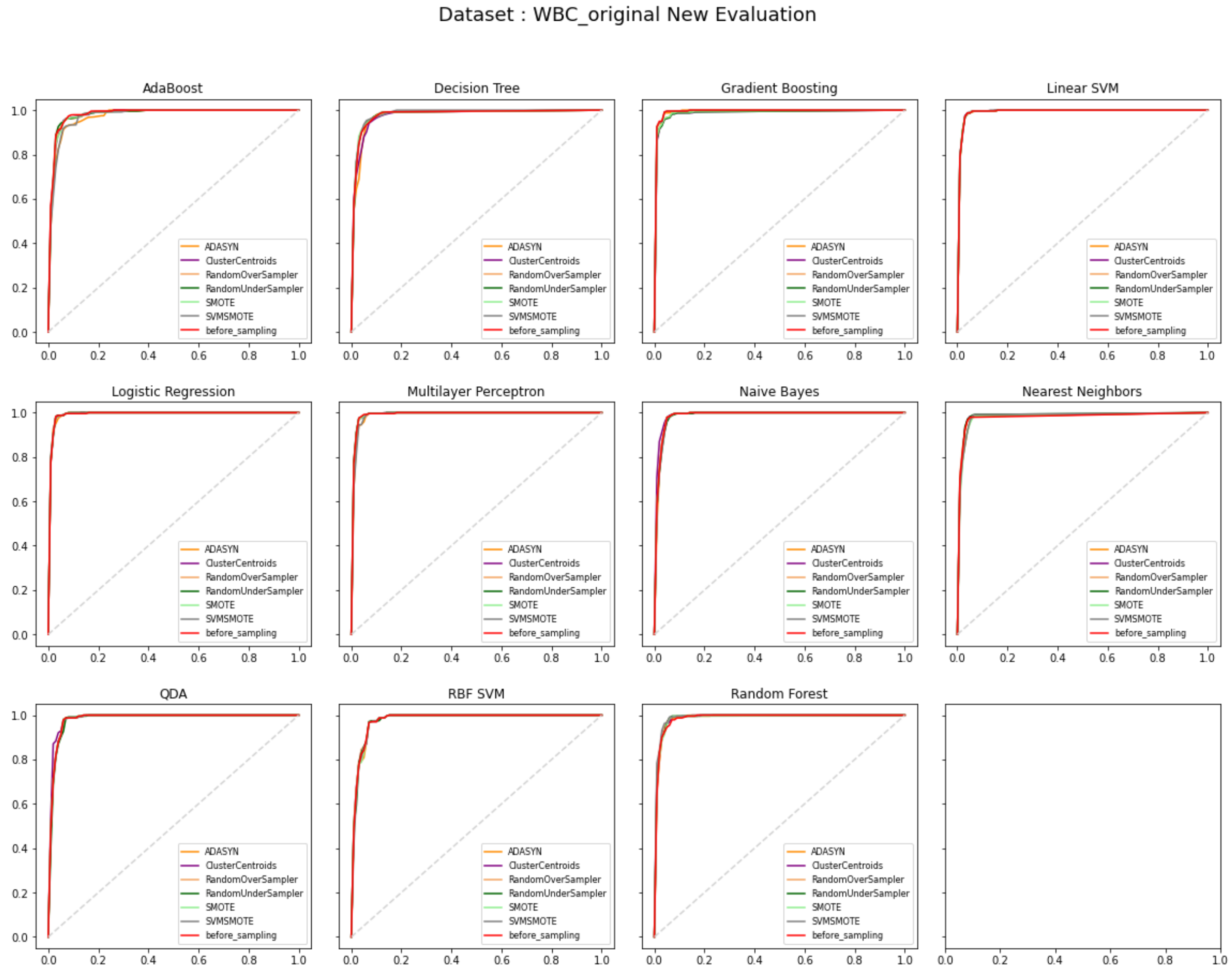

**Appendix Figure  4 WBC Original EFLDA**

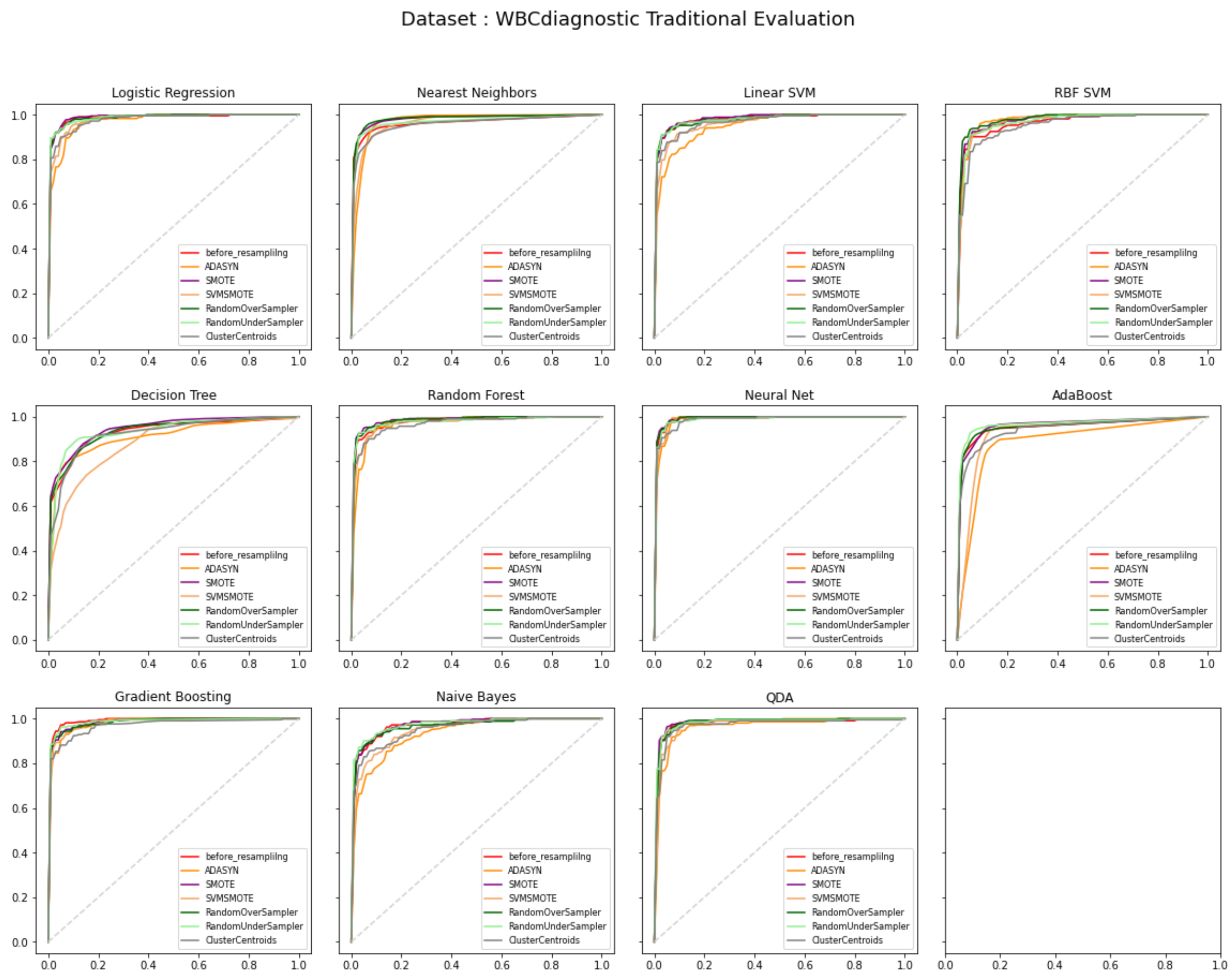

**Appendix Figure 5 WBC Diagnostics -TRA**

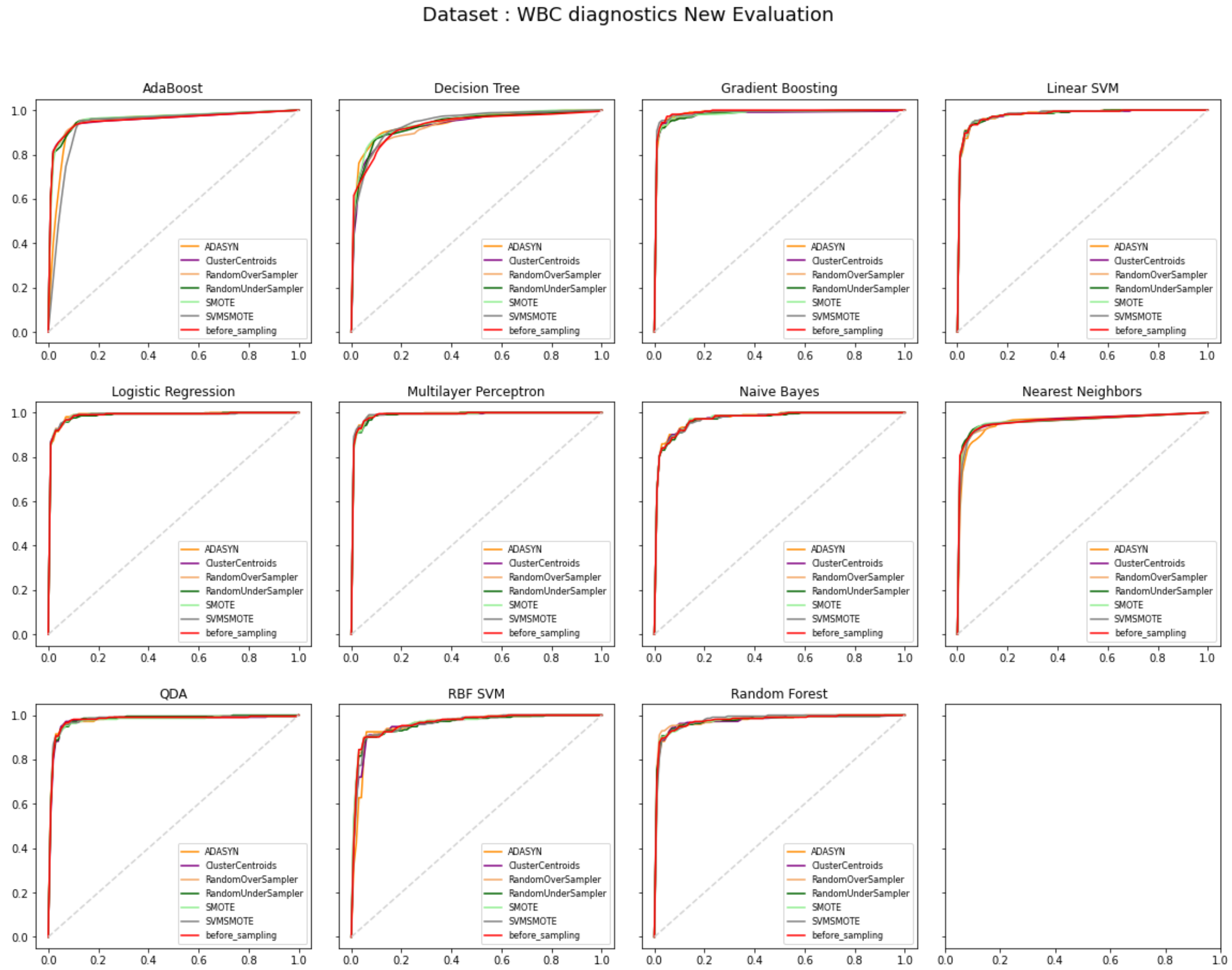

**Appendix Figure 6 WBC Diagnostics -EFLDA**

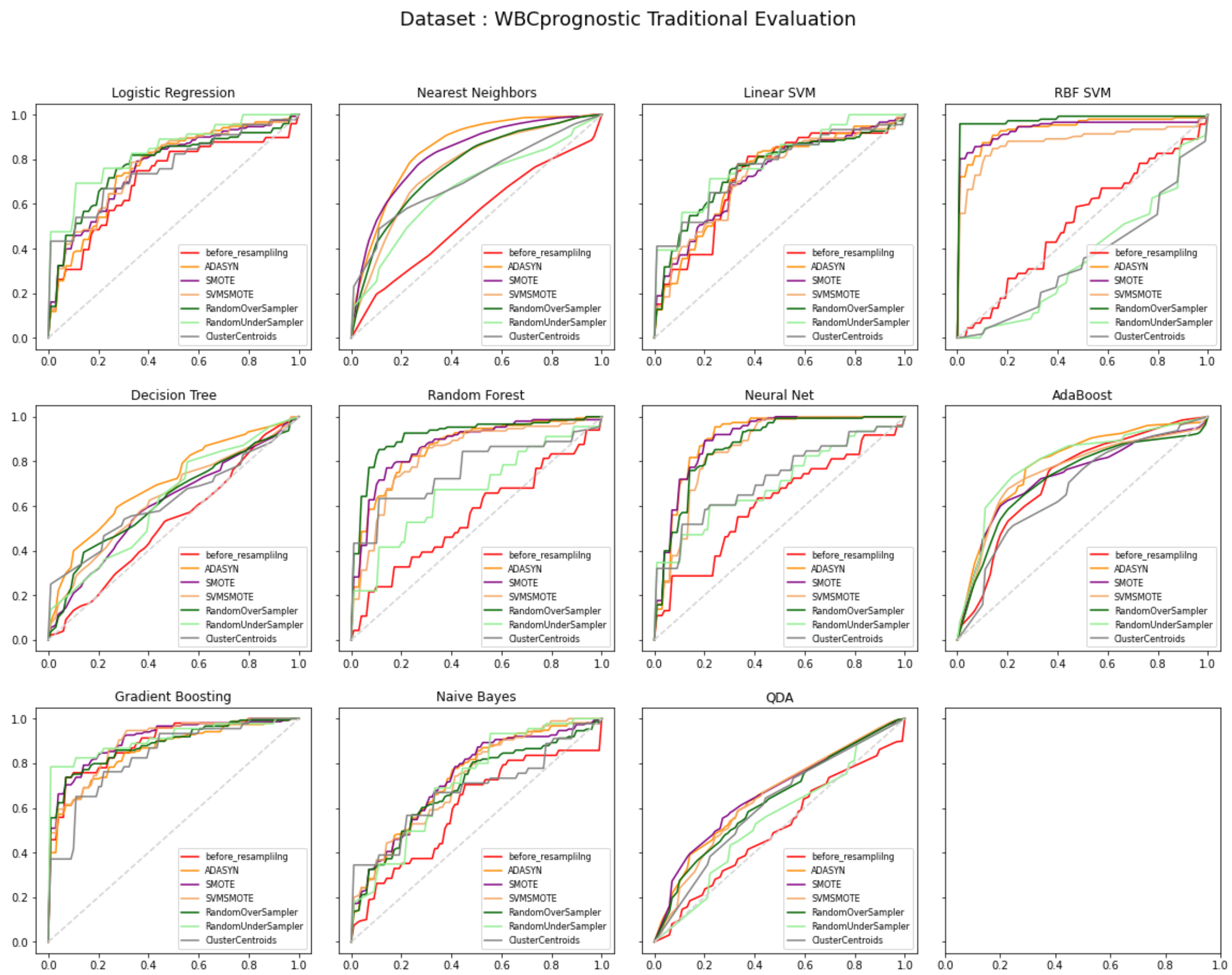

**Appendix Figure 7 WBC Prognostics -TRA**

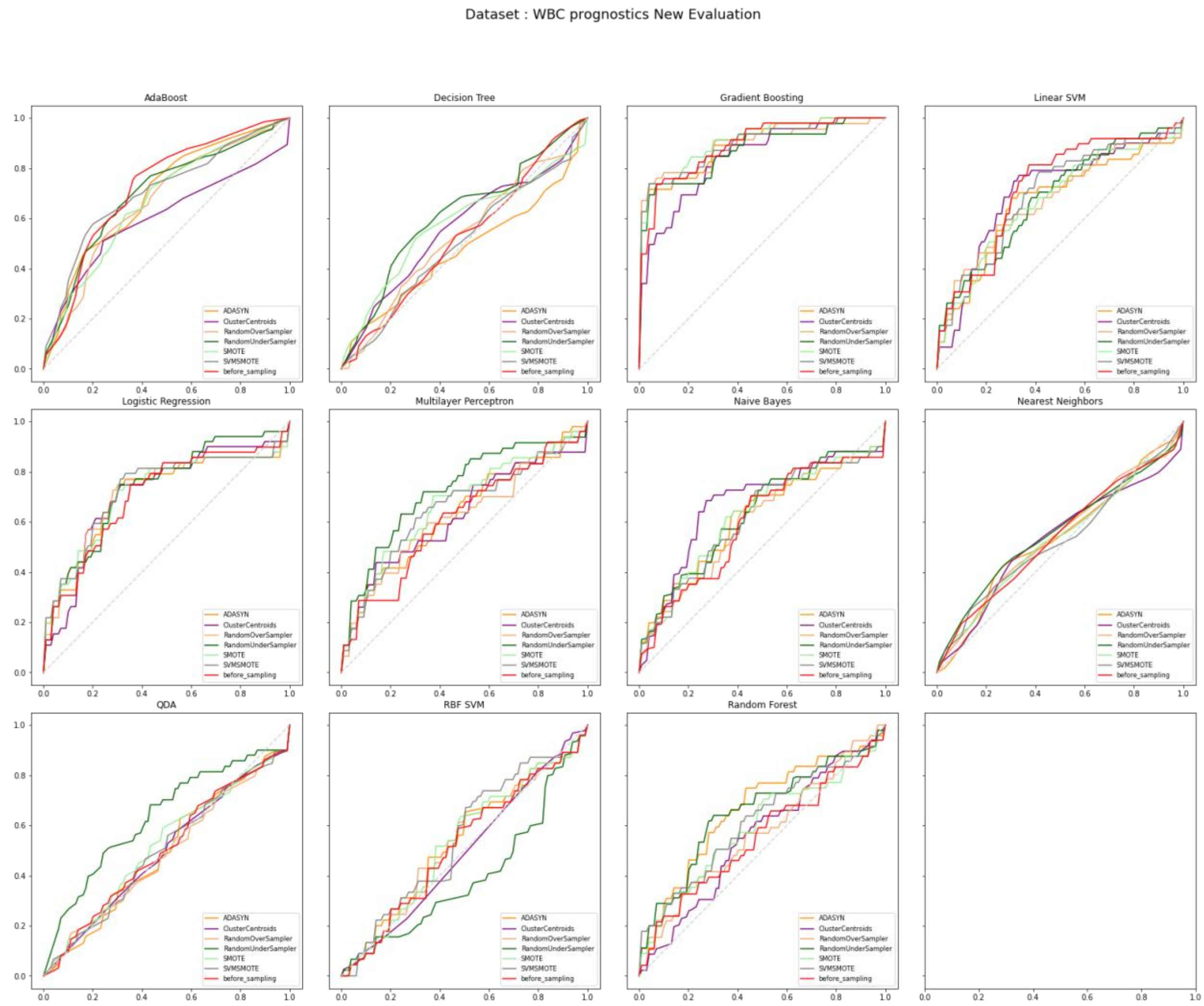

**Appendix Figure 8 WBC Prognostics -EFLDA**

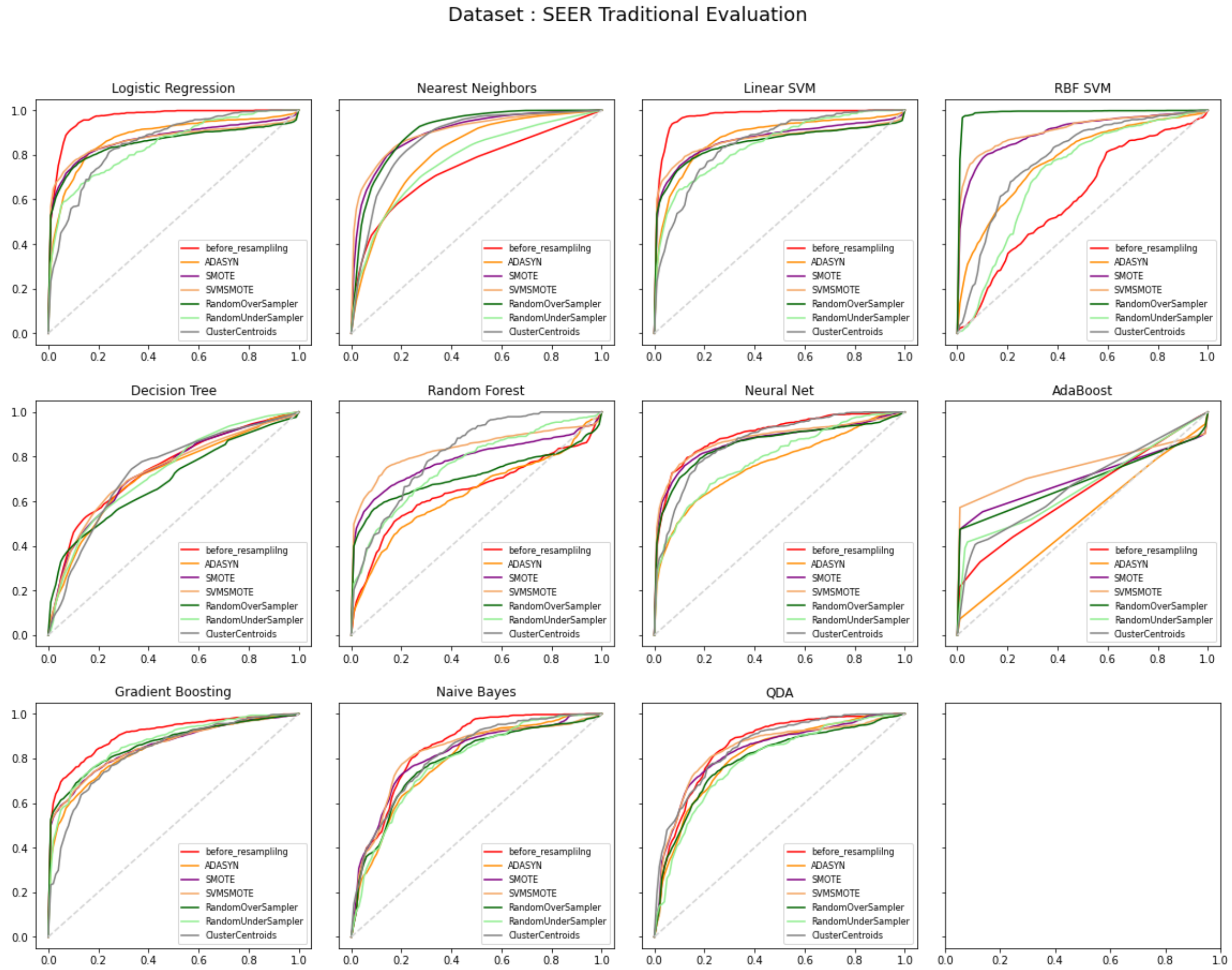

**Appendix Figure 9 SEER- TRA**

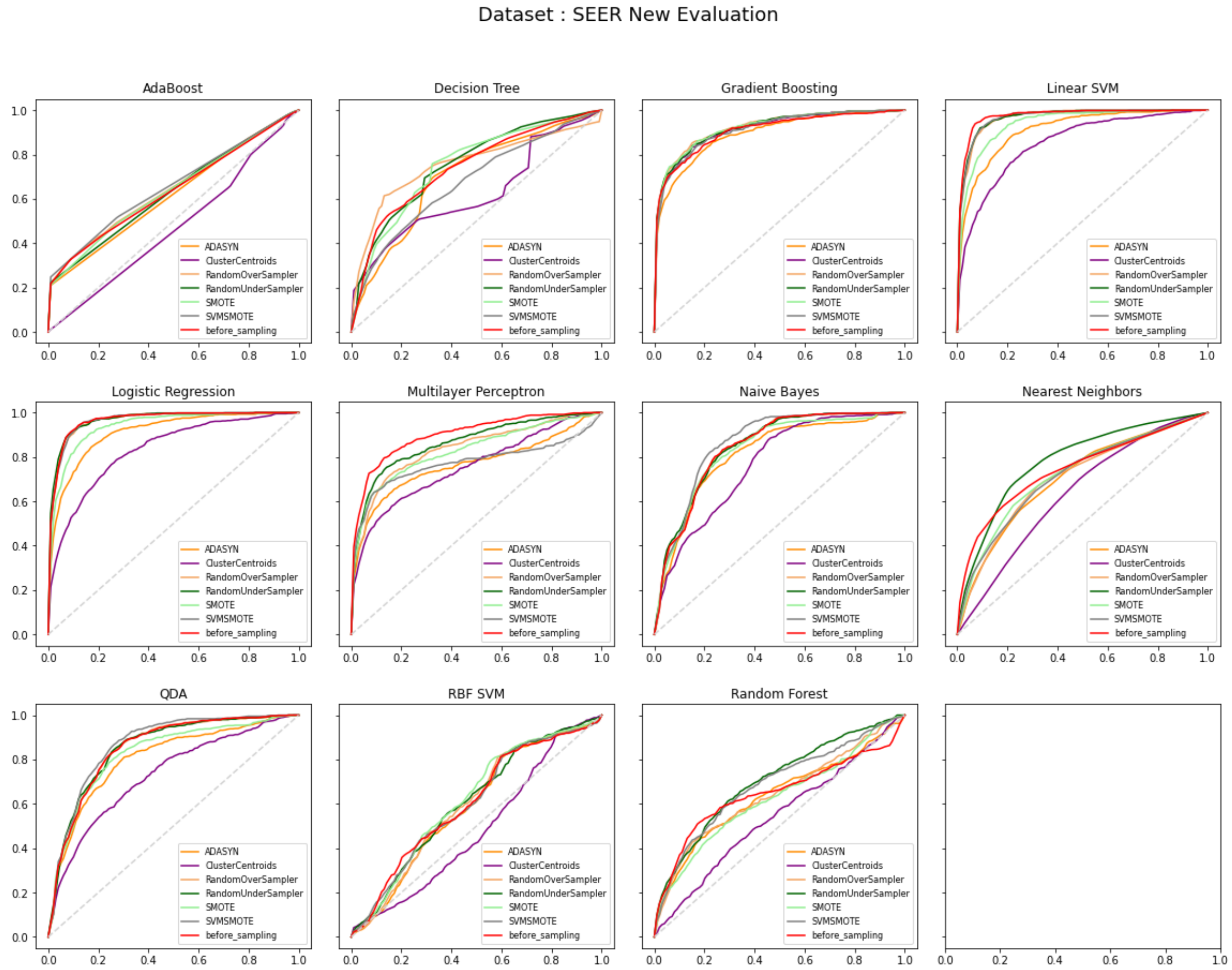

**Appendix Figure 10 SEER-EFLDA**

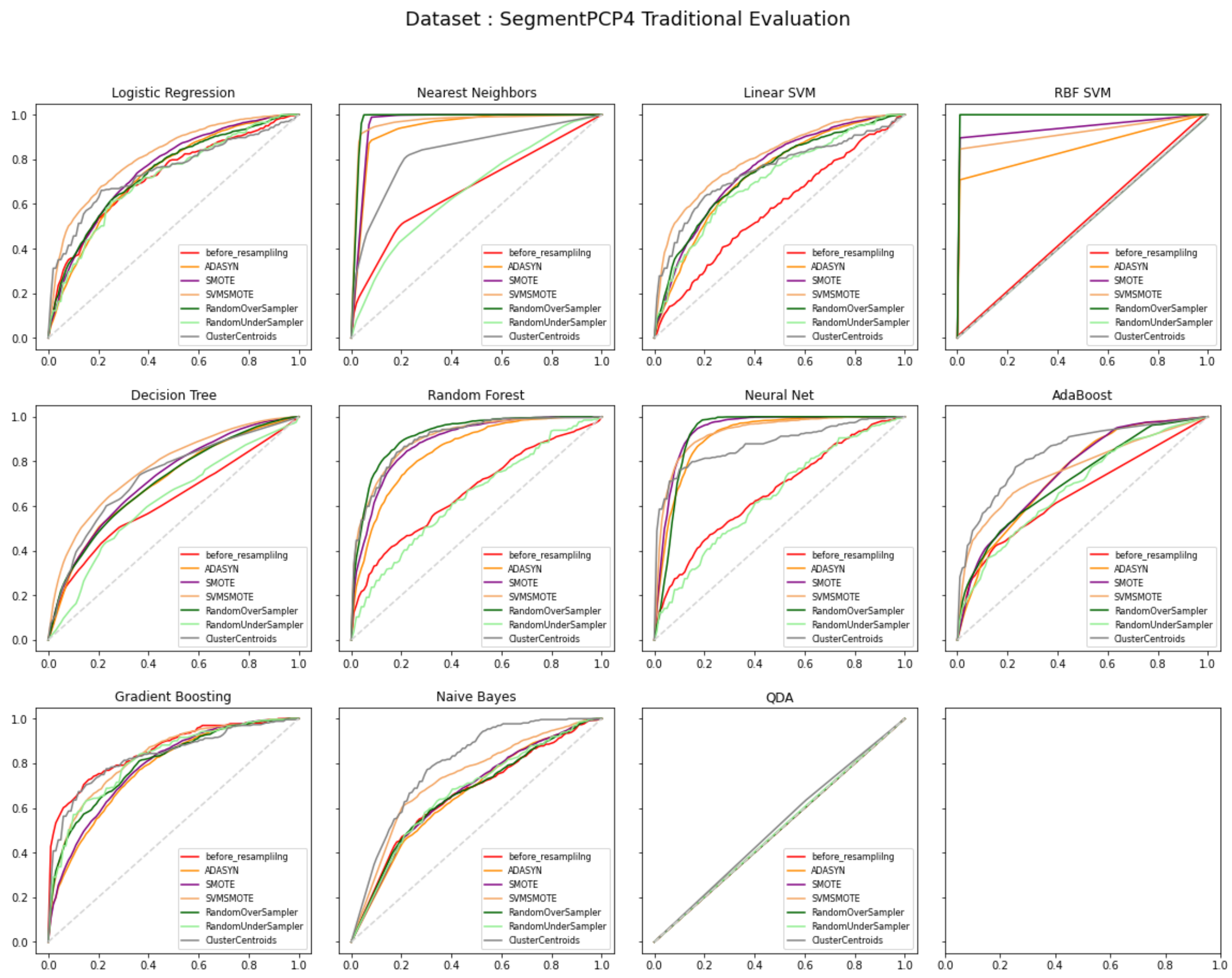

**Appendix Figure 11 SegmentPCP_4 – TRA**

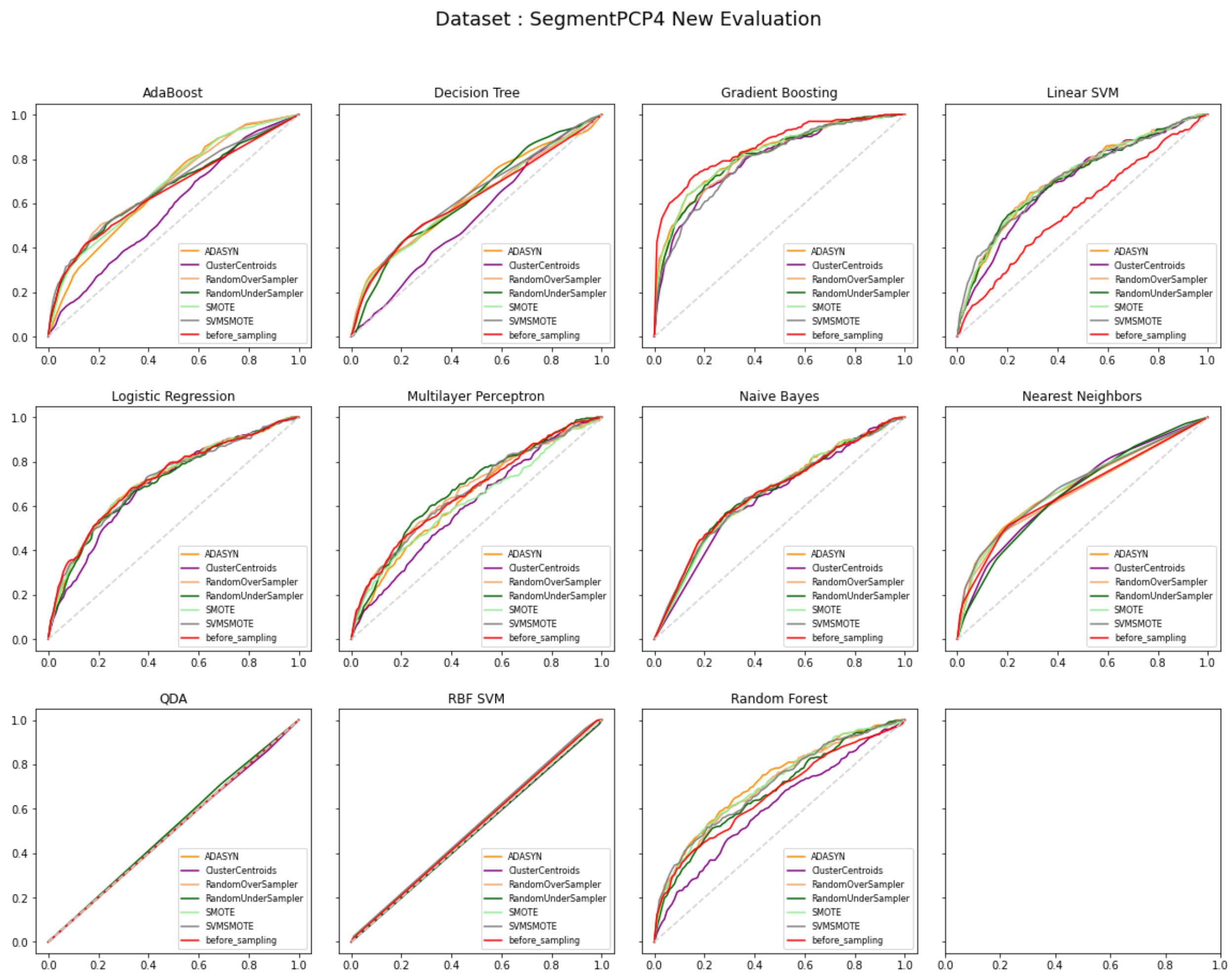

**Appendix Figure 12 SegmentPCP_4 -EFLDA**

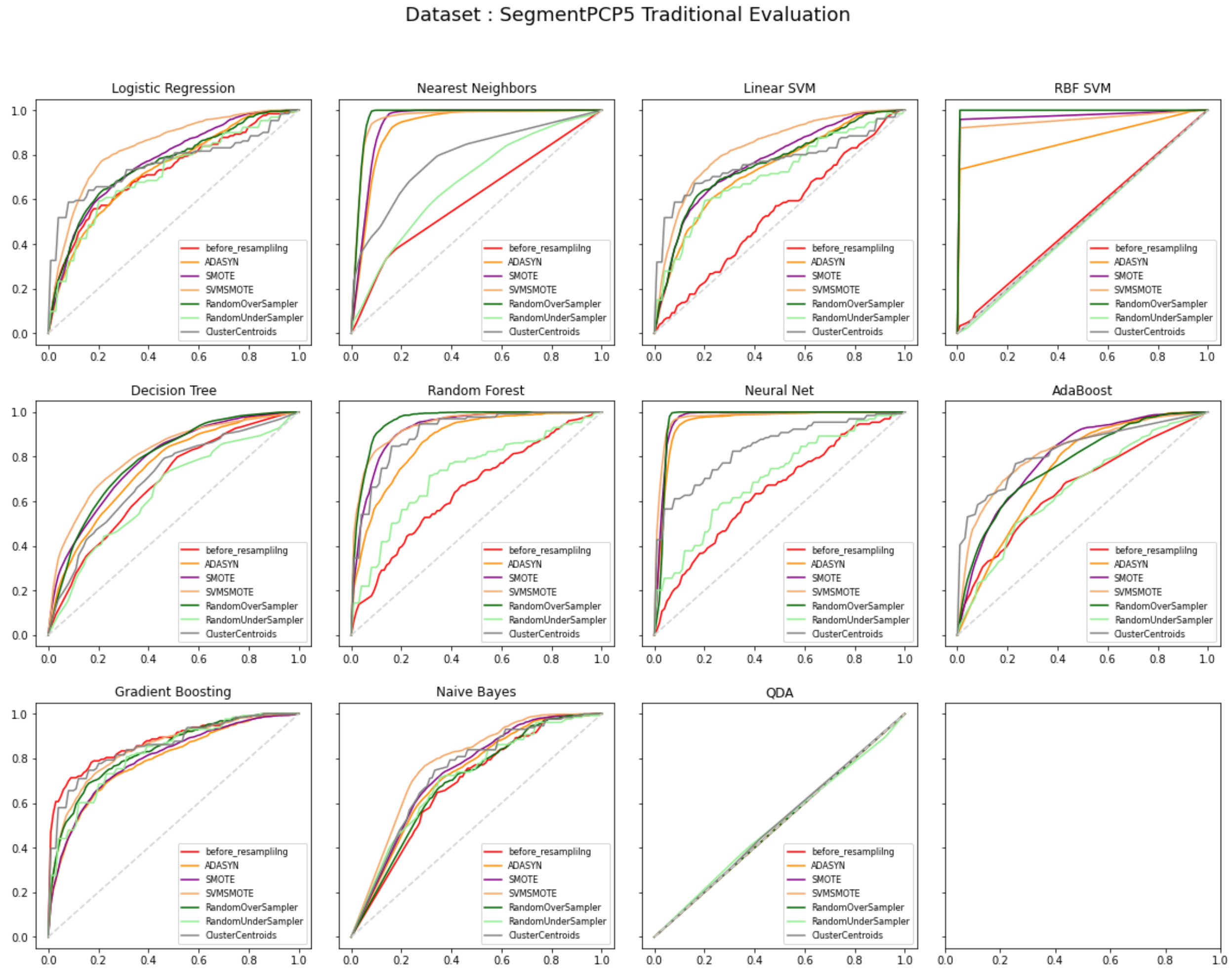

**Appendix Figure 13 SegmentPCP_5 -TRA**

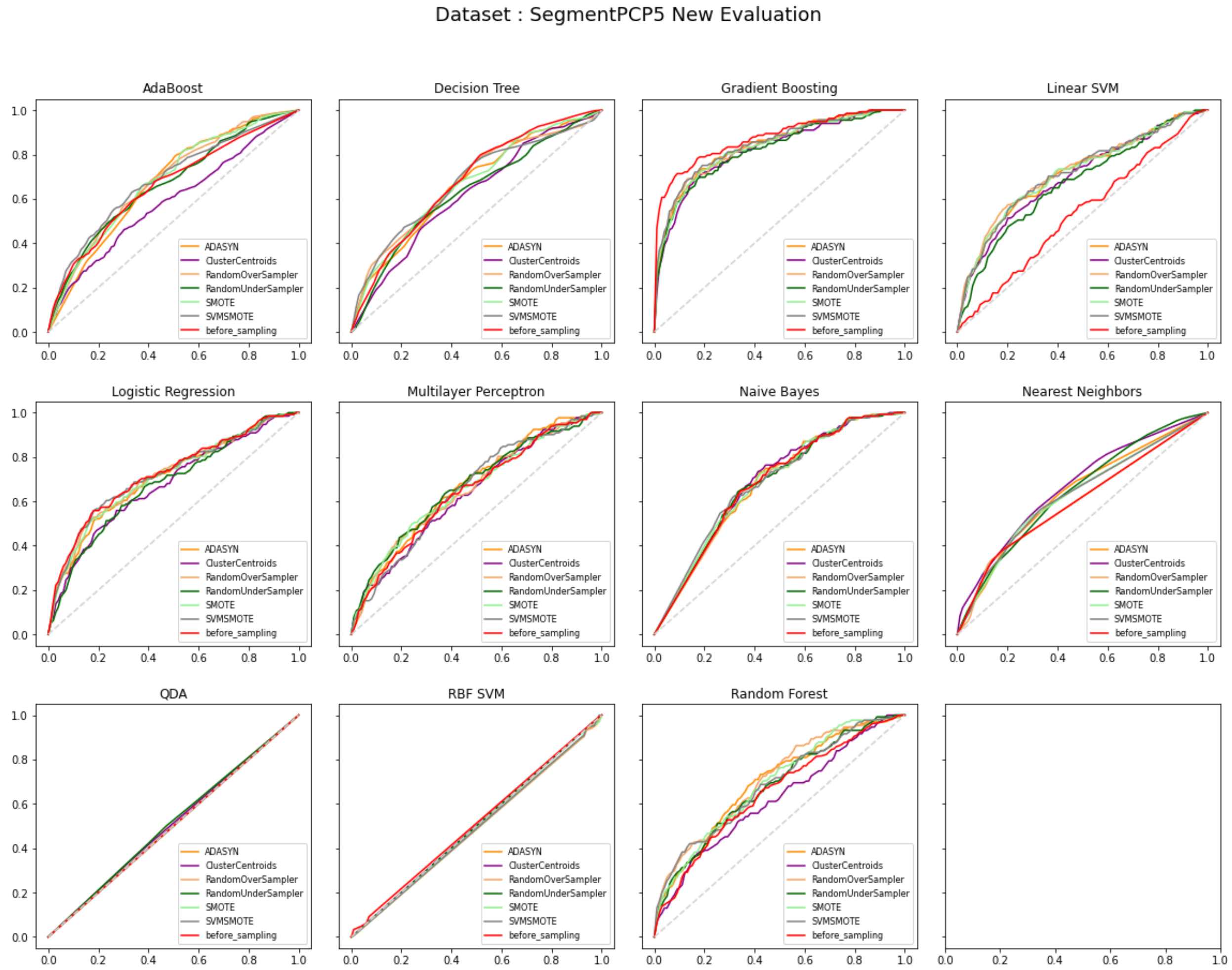

**Appendix Figure 14 SegmentPCP_5 -EFLDA**

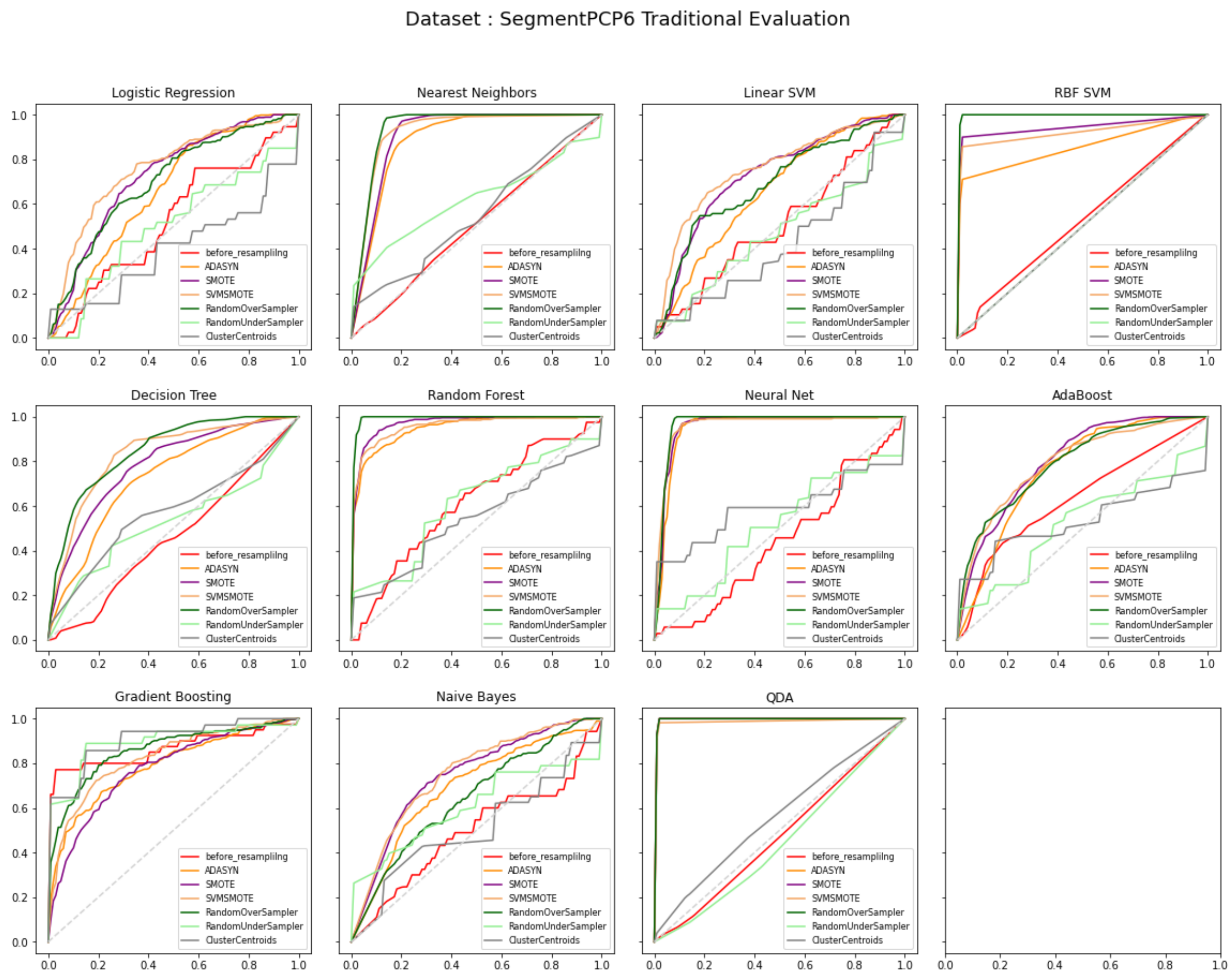

**Appendix Figure 15 SegmentPCP_6 -TRA**

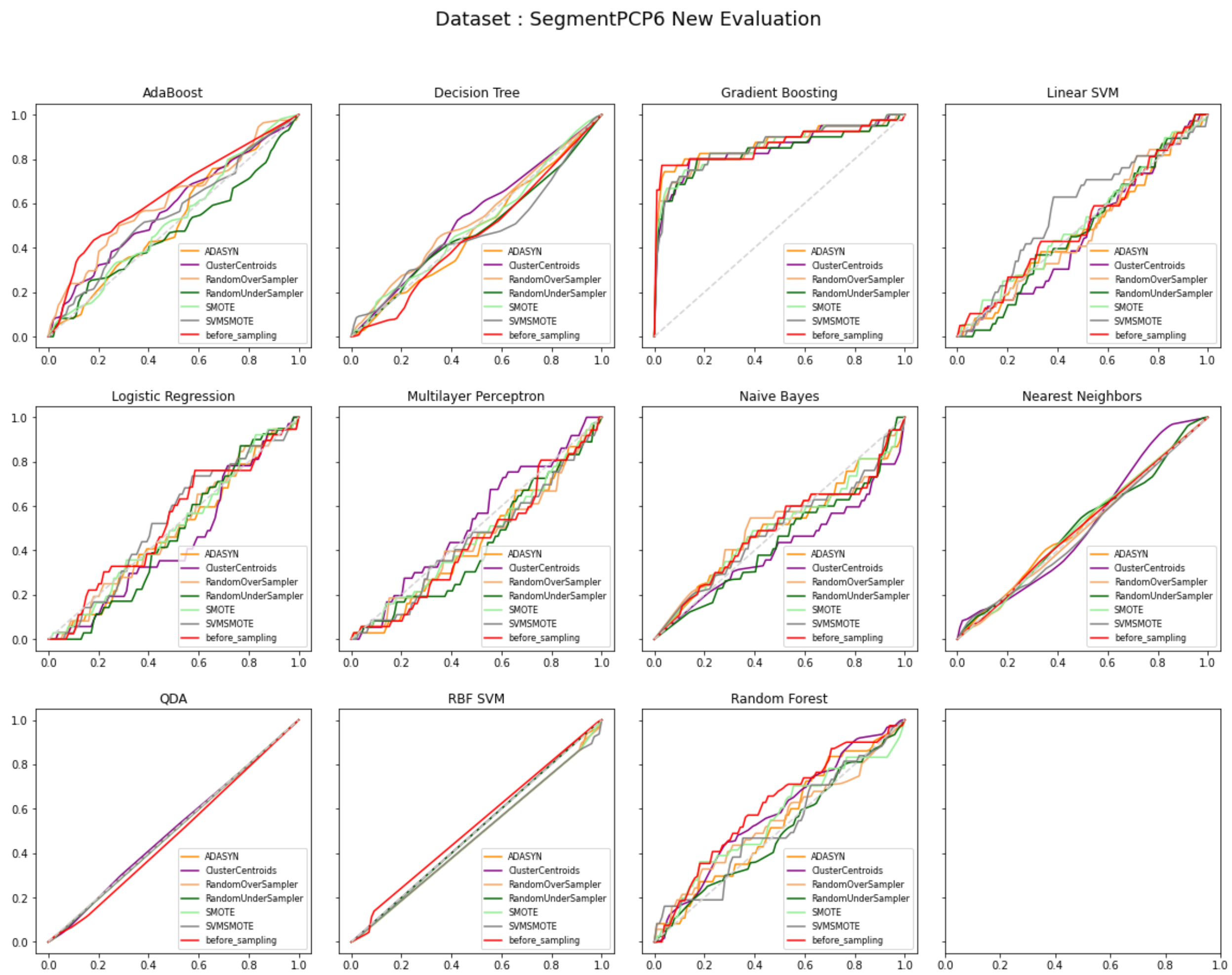

**Appendix Figure 16 SegmentPCP_6-EFLDA**

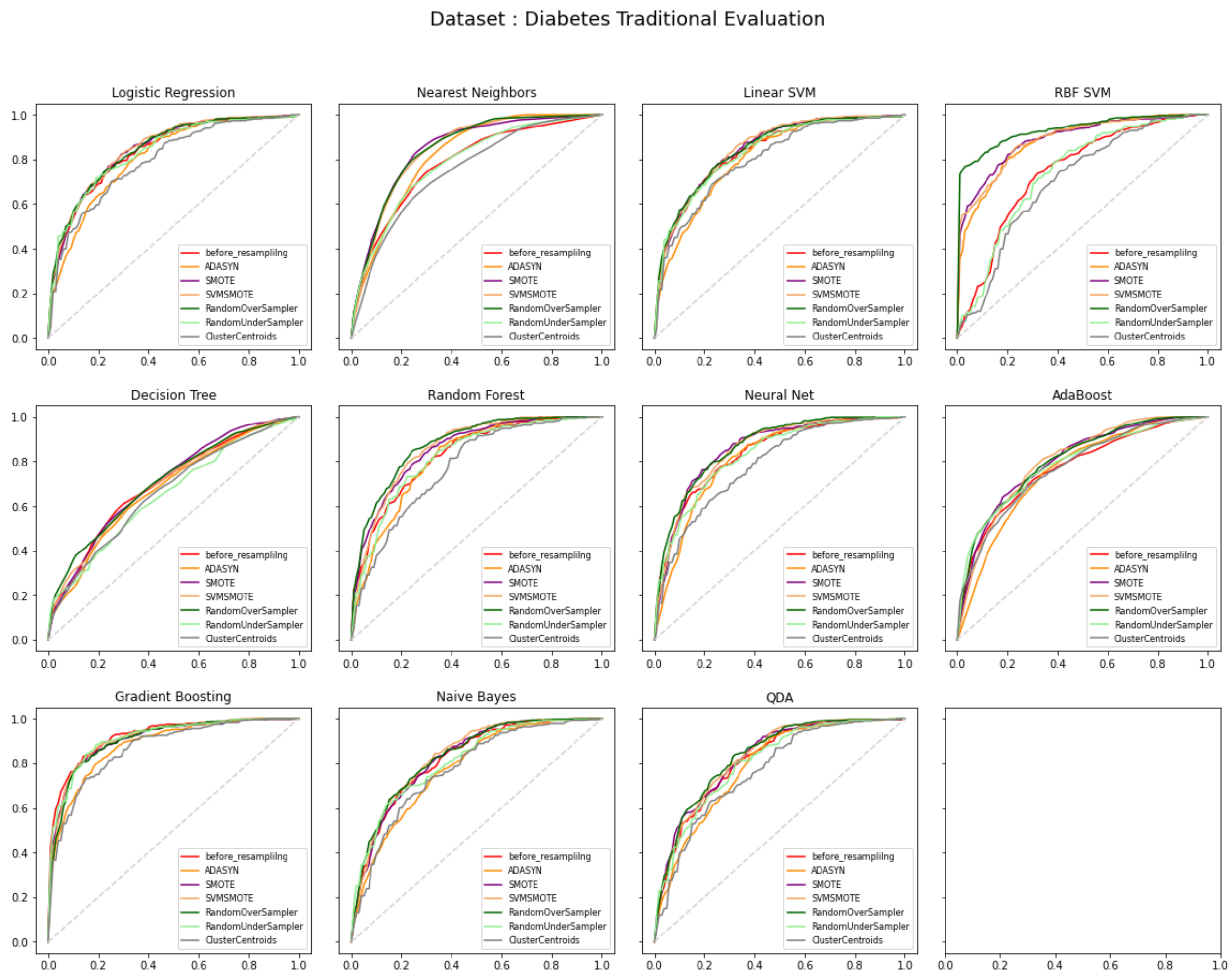

**Appendix Figure 17 Pima_Diabetes -TRA**

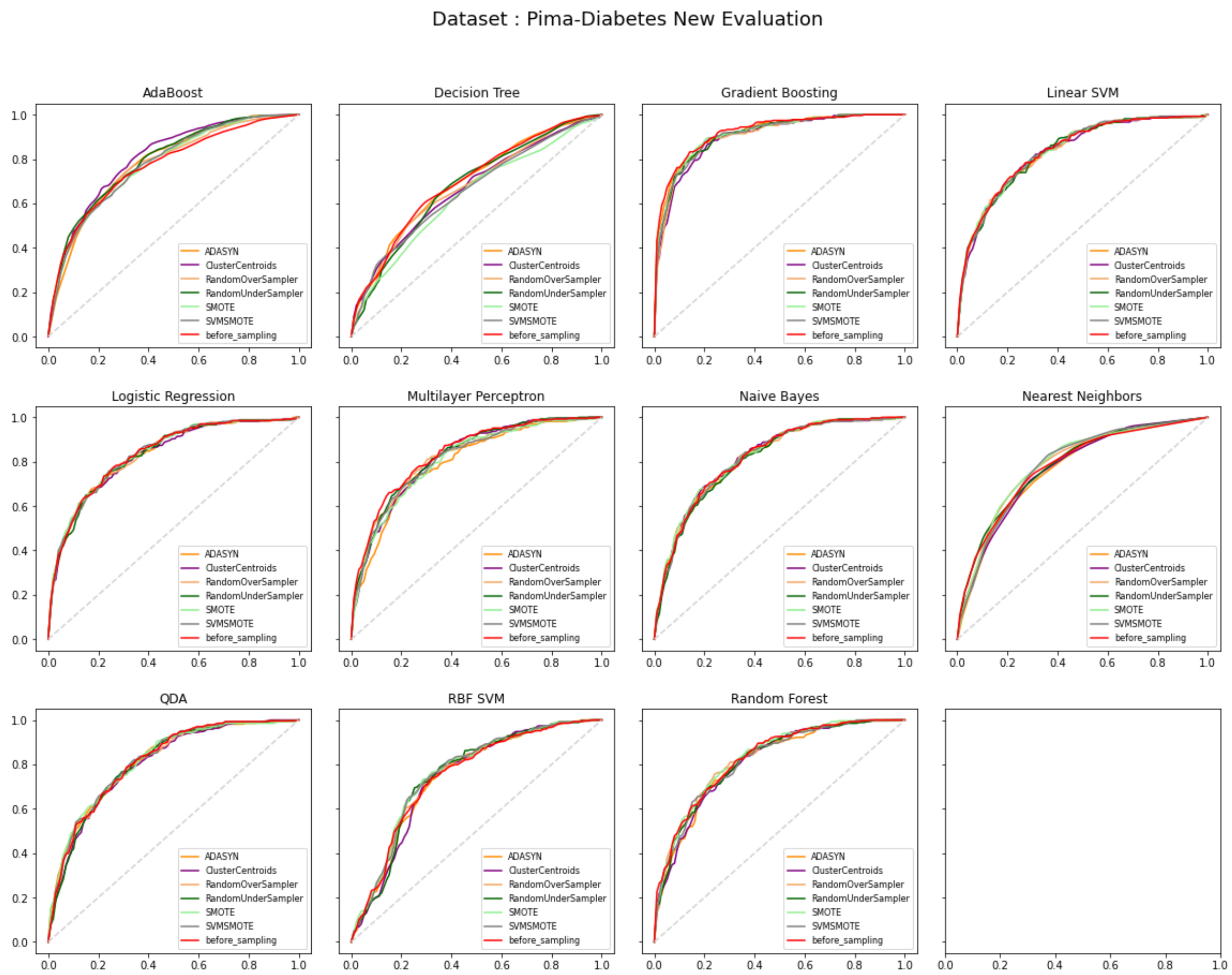

**Appendix Figure 18 Pima_Diabetes - EFLDA**

# Appendix reference: